\documentclass{article}
\usepackage{arxiv}
\usepackage{hyperref}
\usepackage[utf8]{inputenc}
\usepackage{verbatim}
\usepackage{ifthen}
\usepackage{xspace}
\usepackage{graphicx}
\usepackage{pgf}
\usepackage{trimclip}

\usepackage{url}

\usepackage[vlined,ruled]{algorithm2e}
\usepackage{tikz}

\usepackage{color}
\usepackage{xcolor}

\usepackage{ucs}
\usepackage{comment}
\usepackage{amsmath}
\usepackage{fancybox}
\usepackage{caption}
\usepackage{subcaption}
\usepackage{bm}
\usepackage{rotating}

\usepackage{array}
\usepackage{multirow}
\usepackage{colortbl}
\usepackage{xspace}

\newcommand{\burl}[1]{\structure{\url{#1}}}

\newcommand{\ql}{{\sc Q-learning}\xspace}

\newcommand{\dqn}{{\sc dqn}\xspace}
\newcommand{\rainbow}{{\sc rainbow}\xspace}

\newcommand{\awr}{{\sc awr}\xspace}
\newcommand{\ddpg}{{\sc ddpg}\xspace}

\newcommand{\xddpg}{{\sc x-ddpg}\xspace}

\newcommand{\trpo}{{\sc trpo}\xspace}
\newcommand{\ppo}{{\sc ppo}\xspace}

\newcommand{\acer}{{\sc acer}\xspace}

\newcommand{\sac}{{\sc sac}\xspace}

\newcommand{\tddd}{{\sc td}3\xspace}

\newcommand{\aac}{{\sc a}2{\sc c}\xspace}
\newcommand{\aacc}{{\sc aac}\xspace}

\newcommand{\cem}{{\sc cem}\xspace}
\newcommand{\cerl}{{\sc cerl}\xspace}
\newcommand{\erl}{{\sc erl}\xspace}
\newcommand{\epg}{{\sc epg}\xspace}
\newcommand{\pbt}{{\sc pbt}\xspace}
\newcommand{\cemrl}{{\sc cem-rl}\xspace}
\newcommand{\cemacer}{{\sc cem-acer}\xspace}
\newcommand{\arac}{{\sc arac}\xspace} 
\newcommand{\ars}{{\sc ars}\xspace} 
\newcommand{\pssstddd}{{\sc p3s-td3}\xspace} 
\newcommand{\svpg}{{\sc svpg}\xspace} 
\newcommand{\pderl}{{\sc pderl}\xspace}
\newcommand{\pgps}{{\sc pgps}\xspace}
\newcommand{\spderl}{{\sc spderl}\xspace}
\newcommand{\zoac}{{\sc zoac}\xspace}
\newcommand{\derl}{{\sc derl}\xspace}
\newcommand{\deprl}{{\sc deprl}\xspace} 
\newcommand{\serl}{{\sc serl}\xspace} 
 
\newcommand{\superl}{{\sc supe-rl}\xspace} 
\newcommand{\fidirl}{{\sc fidi-rl}\xspace}
\newcommand{\easrl}{{\sc eas-rl}\xspace} 
\newcommand{\scerl}{{\sc sc-erl}\xspace} 
\newcommand{\esac}{{\sc esac}\xspace} 
\newcommand{\evorl}{{\sc evo-RL}\xspace}

\newcommand{\ggn}{{\sc g2n}\xspace}
\newcommand{\ggac}{{\sc g2ac}\xspace}
\newcommand{\ggppo}{{\sc g2ppo}\xspace}
\newcommand{\grac}{{\sc grac}\xspace}
\newcommand{\searl}{{\sc searl}\xspace}

\newcommand{\eQ}{{\sc eQ}\xspace}
\newcommand{\gadrl}{{\sc ga-drl}\xspace}
\newcommand{\ohtes}{{\sc oht-es}\xspace}
\newcommand{\nsrl}{{\sc ns-rl}\xspace}
\newcommand{\pnsrl}{{\sc pns-rl}\xspace}
\newcommand{\cmamega}{{\sc cma-mega}\xspace}

\newcommand{\cmaes}{{\sc cma-es}\xspace}

\newcommand{\her}{\textsc{her}\xspace}

\newcommand{\sges}{{\sc sges}\xspace}

\newcommand{\qtopt}{{\sc qt-opt}\xspace}
\newcommand{\saccepo}{{\sc sac-cepo}\xspace}
\newcommand{\cgp}{{\sc cgp}\xspace}
\newcommand{\tres}{{\sc tres}\xspace}
\newcommand{\nes}{{\sc nes}\xspace}
\newcommand{\dvd}{{\sc d}v{\sc d}\xspace}

\newcommand{\pets}{{\sc pets}\xspace}
\newcommand{\poplin}{{\sc poplin}\xspace}
\newcommand{\planet}{{\sc plaNet}\xspace}
\newcommand{\bnet}{{\sc bnet}\xspace}

\newcommand{\zospi}{{\sc zospi}\xspace}
\newcommand{\pso}{{\sc pso}\xspace}
\newcommand{\cspc}{{\sc cspc}\xspace}
\newcommand{\chdrl}{{\sc chdrl}\xspace}

\newcommand{\geppg}{{\sc gep-pg}\xspace}

\newcommand{\qdrl}{{\sc qd-rl}\xspace}
\newcommand{\qdpg}{{\sc qd-pg}\xspace}

\newcommand{\pgame}{{\sc pga-me}\xspace}

\newcommand{\nsres}{{\sc nsr-es}\xspace}

\newcommand{\argmax}{\mathop{\rm argmax}}

\newcommand{\rien}[1]{}

\DeclareGraphicsExtensions{.jpg,.mps,.pdf,.png}

\usepackage{colortbl}

\definecolor{myred}{rgb}{0.8,0,0}
\definecolor{mygreen}{rgb}{0,0.6,0}
\definecolor{myblue}{rgb}{0,0,0.7}

\newcounter{ques} 
\newcommand{\ques}{\arabic{ques}}

\newcommand\wi[1]{$\circ$}
\newcommand\bu[1]{$\bullet$}
\newcommand\ot[1]{$\star$}
\newcommand\bo[1]{$\bullet\star$}

\newcommand\pr[1]{\cellcolor{orange!30} $\circ$}
\newcommand\ye[1]{\cellcolor{green!30} $\bullet$}
\newcommand\no[1]{\cellcolor{red!30}{\bf x}}
\newcommand\na[1]{\cellcolor{blue!30} $\spadesuit$}

\newcommand{\shortrightarrow}{\clipbox*{{.25\width} 0pt {\width} {\height}} \textrightarrow}

\usepackage[english]{babel}

\usepackage{graphicx}
\usepackage{subfloat}
\usepackage{array, makecell}
\usepackage{booktabs,multirow}
\usepackage{soul}
\usepackage{diagbox}

\begin{document}

\title{Combining Evolution and Deep Reinforcement Learning for Policy Search: a Survey}

\author{\href{https://orcid.org/0000-0002-8544-0229}{\includegraphics[scale=0.06]{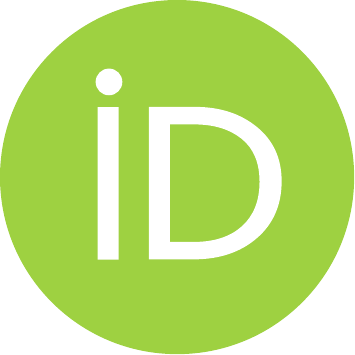}\hspace{1mm}Olivier Sigaud},
\\
Sorbonne Universit\'e, CNRS, Institut des Syst\`emes Intelligents et de Robotique,\\ F-75005 Paris, France\\
\tt{Olivier.Sigaud@isir.upmc.fr}}

\maketitle

\begin{abstract}
Deep neuroevolution and deep Reinforcement Learning have received a lot of attention in the last years. Some works have compared them, highlighting theirs pros and cons, but an emerging trend consists in combining them so as to benefit from the best of both worlds. In this paper, we provide a survey of this emerging trend by organizing the literature into related groups of works and casting all the existing combinations in each group into a generic framework.
We systematically cover all easily available papers irrespective of their publication status, focusing on the combination mechanisms rather than on the experimental results. In total, we cover 45 algorithms more recent than 2017. We hope this effort will favor the growth of the domain by facilitating the understanding of the relationships between the methods, leading to deeper analyses, outlining missing useful comparisons and suggesting new combinations of mechanisms.
\end{abstract}
\maketitle

\section{Introduction}

The idea that the extraordinary adaptive capabilities of living species results from a combination of evolutionary mechanisms acting at the level of a population and learning mechanisms acting at the level of individuals is ancient in life sciences \citep{simpson1953baldwin} and has inspired early work in Artificial Intelligence (AI) research \citep{holland1978cognitive}. This early starting point has led to the independent growth of two bodies of formal frameworks, evolutionary methods and reinforcement learning (RL). The early history of the evolutionary side is well covered in \cite{back1997evolutionary} and from the RL side in \cite{sutton2018reinforcement}. Despite these independent developments, research dedicated to the combination has remained active, in particular around Learning Classifier Systems \citep{lanzi1999analysis, sigaud2007learning} and studies of the Baldwin effect \citep{weber2003evolution}.
A broader perspective and survey on all the evolutionary and RL combinations anterior to the advent of the so called ''deep learning" methods using large neural networks can be found in \cite{drugan2019reinforcement}.

In this paper, we propose a survey of a renewed approach to this combination that builds on the unprecedented progress made possible in evolutionary and deep RL methods by the growth of computational power and the availability of efficient libraries to use deep neural networks. As this survey shows, the topic is rapidly gaining popularity with a wide variety of approaches and even emerging libraries dedicated to their implementation \citep{tangri2022pearl}. Thus we believe it is the right time for laying solid foundations to this growing field, by listing the approaches and providing a unified view that encompasses them. There are recent surveys about the comparison of evolutionary and RL methods \citep{qian2021derivative, majid2021deep} which mention the emergence of some of these combinations. With respect to these surveys, ours is strictly focused on the combinations and attempts to provide a list of relevant papers as exhaustive as possible at the time of its publication, irrespective of their publication status.

This survey is organized into groups of algorithms using the evolutionary part for the same purpose. In Section~\ref{sec:policies}, we first review algorithms where evolution is looking for efficient policies, that is combining deep neuroevolution and deep RL. We then cover in Section~\ref{sec:actions} algorithms where evolution directly looks for efficient actions in a given state rather than for policies. In Section~\ref{sec:diversity}, we cover the combination of deep RL algorithm with diversity seeking methods. Finally, in Section~\ref{sec:other}, we cover various other uses of evolutionary methods, such as optimizing hyperparameters or the system's morphology.
To keep the survey as short as possible, we consider that the reader is familiar with evolutionary and RL methods in the context of policy search, and has a good understanding of their respective advantages and weaknesses. We refer the reader to \cite{sigaud2019policy} for an introduction of the methods and to surveys about comparisons to know more about their pros and cons \citep{qian2021derivative, majid2021deep}.

\section{Evolution of policies for performance}
\label{sec:policies}

\begin{figure}[!htbp]
  \begin{center}
{\includegraphics[width=0.4\linewidth]{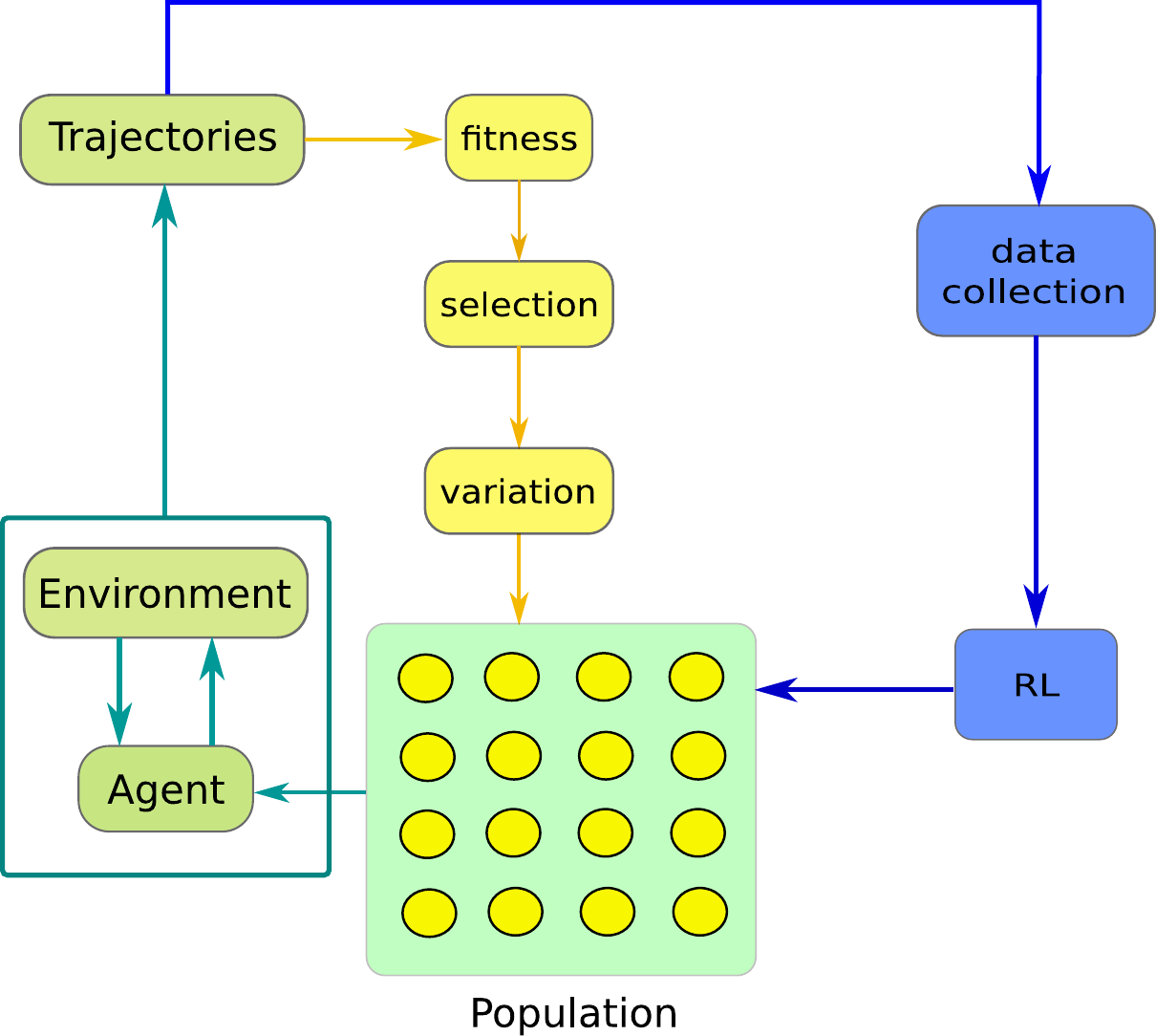}}
   \caption{
   The general template of algorithms combining deep neuroevolution and deep RL. A population of agents interact with an environment, and produce trajectories composed of states, actions and rewards. From the left-hand side, an evolutionary loop selects and evolves these agents based on their fitness, which is computed holistically over trajectories. From the right-hand side, a deep RL loop improves one or several agents using a gradient computed over the elementary steps of trajectories stored into a replay buffer. \label{fig:generic}}
   \end{center}
\end{figure}

The methods in our first family combine a deep neuroevolution loop and a deep RL loop. Figure~\ref{fig:generic} provides a generic template to illustrate such combinations. The central question left open by the template is how both loops interact with each other. Note that this template is not much adapted to account for works where the combination is purely sequential, such as \cite{kim2007hybrid} or the \geppg algorithm \citep{colas2018gep}. Besides, to avoid any confusion with the multi-agent setting, note that agents are interacting in isolation with their own copy of the environment and cannot interact with each other.

\begin{table*}[htbp]
\centering
\caption{Combinations evolving policies for performance. The table states whether the algorithms in the rows use the mechanisms in the columns. The colors are as follows. In the column about other combination mechanisms (+ Comb. Mech.): Critic gradient addition \ye{1} (green), Population from Actor \na{1} (blue), None \no{1} (red). In all other columns: \ye{1} (green): yes, \no{1} (red): no.
In \bnet, BBNE stands for Behavior-Based NeuroEvolution and CPG stands for Cartesian Genetic Programming \citep{miller2009cartesian}. The different GA labels stand for various genetic algorithms, we do not go into the details.
\label{tab:algos1}}
\resizebox{\textwidth}{!}{\begin{tabular}{|r|r|r|r|r|r|r|r|}
    \hline
    \backslashbox{Algo.}{Prop.} &\makecell{RL\\ algo.} & \makecell{Evo.\\ algo.} & \makecell{Actor\\ Injec.} & \makecell{+ Comb.\\ Mech.} & \makecell{Surr.\\ Fitness} & \makecell{Soft\\ Update} & \makecell{Buffer\\ Filt.} \\  \hline
    \multicolumn{1}{|l|}{\erl \cite{khadka2018evolutionaryNIPS}} & \ddpg & GA & \ye{1} & \no{1} & \no{1} & \no{1} & \no{1}   \\ \hline
    \multicolumn{1}{|l|}{\cerl \cite{khadka2019collaborative}} & \tddd& GA & \ye{1} & \no{1} & \no{1} & \no{1} & \no{1} \\ \hline
    \multicolumn{1}{|l|}{\pderl \cite{bodnar2020proximal}} & \tddd & GA  & \ye{1} & \no{1} & \no{1} & \no{1} & \no{1} \\ \hline
    \multicolumn{1}{|l|}{\esac \cite{suri2020maximum}} & \sac & ES  & \ye{1} & \no{1} & \no{1} & \no{1} & \no{1} \\ \hline
    \multicolumn{1}{|l|}{\fidirl \cite{shi2019fidi}} & \ddpg & \ars  & \ye{1} & \no{1} & \no{1} & \no{1} & \no{1} \\ \hline
    \multicolumn{1}{|l|}{\xddpg \cite{espositi2020gradient}} & \ddpg & GA  & \ye{1} & \no{1} & \no{1} & \no{1} & \no{1} \\ \hline
    \multicolumn{1}{|l|}{\cemrl \cite{pourchot2018cem}} & \tddd & \cem & \no{1} & \ye{1} & \no{1} & \no{1} & \no{1}   \\ \hline
    \multicolumn{1}{|l|}{\cemacer \cite{tang2021guiding}} & \acer & \cem & \no{1} & \ye{1} & \no{1} & \no{1} & \no{1}   \\ \hline
    \multicolumn{1}{|l|}{\serl \cite{wang2022surrogate}} & \ddpg & GA & \ye{1} & \no{1} & \ye{1} & \no{1} & \no{1}   \\ \hline
    \multicolumn{1}{|l|}{\spderl \cite{wang2022surrogate}} & \tddd & GA & \ye{1} & \no{1} & \ye{1}  & \no{1}  & \no{1}  \\ \hline
    \multicolumn{1}{|l|}{\pgps \cite{kim2020pgps}} & \tddd & \cem & \ye{1} & \no{1} & \ye{1} & \ye{1}& \no{1} \\ \hline
    \multicolumn{1}{|l|}{\bnet \cite{stork2021behavior}} & BBNE & CPG & \ye{1} & \no{1} & \ye{1} & \no{1} & \no{1} \\ \hline
    \multicolumn{1}{|l|}{\cspc \cite{zheng2020cooperative}} & \sac + \ppo & \cem & \ye{1} & \no{1} & \no{1} & \no{1} & \ye{1} \\ \hline
    \multicolumn{1}{|l|}{\superl \cite{marchesini2021genetic}} & \rainbow or \ppo & GA & \ye{1} & \na{1} & \no{1} & \ye{1} & \no{1} \\ \hline
    \multicolumn{1}{|l|}{\ggac \cite{chang2018genetic}} & \aac  & GA & \no{1} & \na{1} & \no{1} & \no{1} & \no{1} \\ \hline
    \multicolumn{1}{|l|}{\ggppo \cite{chang2018genetic}} &  \ppo & GA & \no{1} & \na{1} & \no{1} & \no{1} & \no{1} \\ \hline
\end{tabular}}
\end{table*}


The main motivation for combining evolution and deep RL is the improved performance that may result from the combination. For instance, through simple experiments with simple fitness landscapes and simplified versions of the components, combining evolution and RL can be shown to work better than using either of the two in isolation \citep{todd2020interaction}. Why is this so? One of the explanations is the following. A weakness of policy gradient methods at the heart of deep RL is that they compute an estimate of the true gradient based on a limited set of samples. This gradient can be quite wrong due to the high variance of the estimation, but it is applied blindly to the current policy without checking that this actually improves it. By contrast, variation-selection methods at the heart of evolutionary methods evaluate all the policies they generate and remove the poorly performing ones. Thus a first good reason to combine policy gradient and variation-selection methods is that the latter may remove policies that have been deteriorated by the gradient step. Below we list different approaches building on this idea.
This perspective is the one that gives rise to the largest list of combinations. We further split this list into several groups of works in the following sections.

\subsection{Deep RL actor injection}
\label{sec:dai}

\begin{figure}[!htbp]
  \begin{center}
  \subfloat[\label{fig:erl_cerl}]{\includegraphics[width=0.36\linewidth]{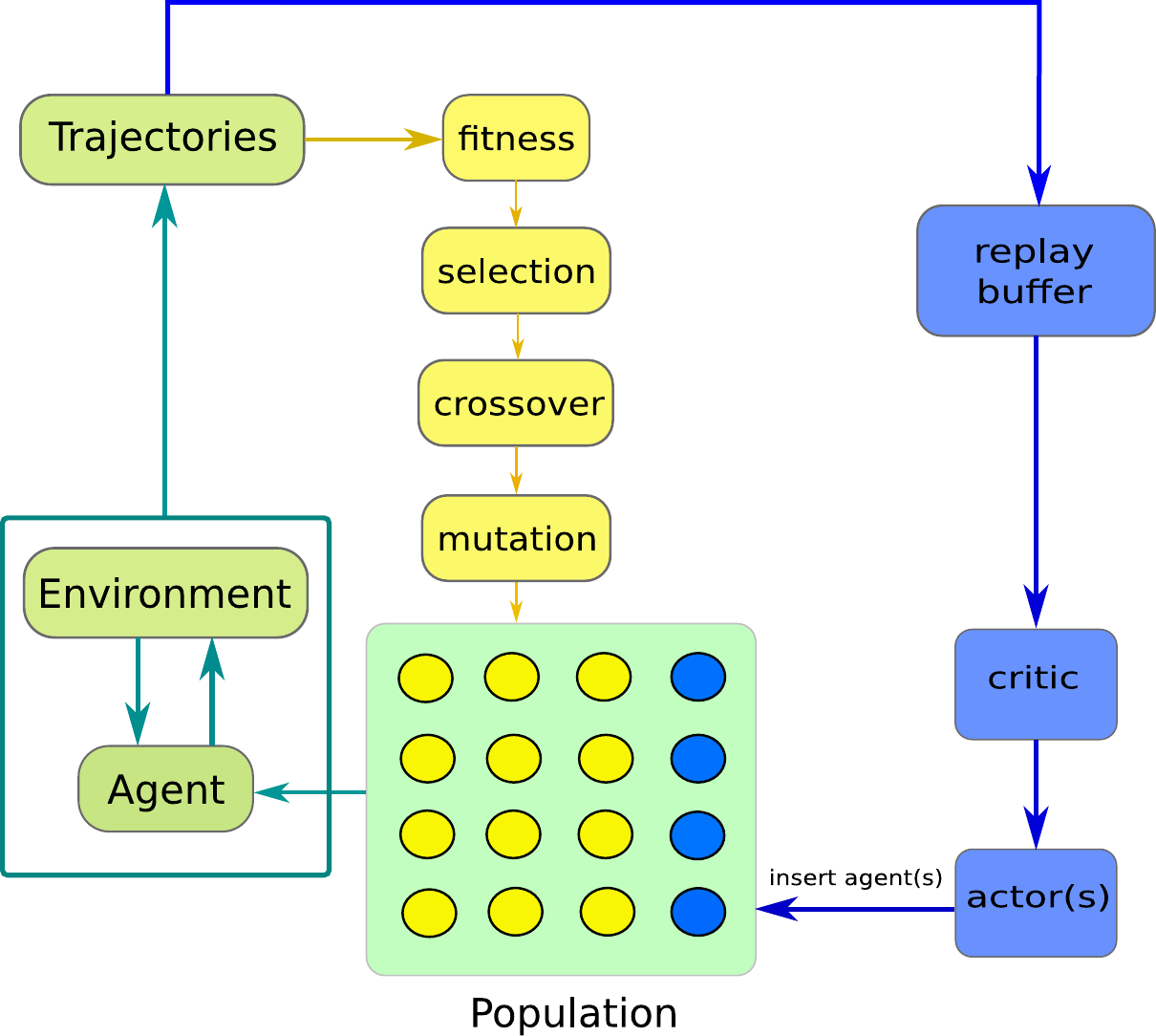}}
  \hspace{0.7cm}
  \subfloat[\label{fig:pderl}]{\includegraphics[width=0.36\linewidth]{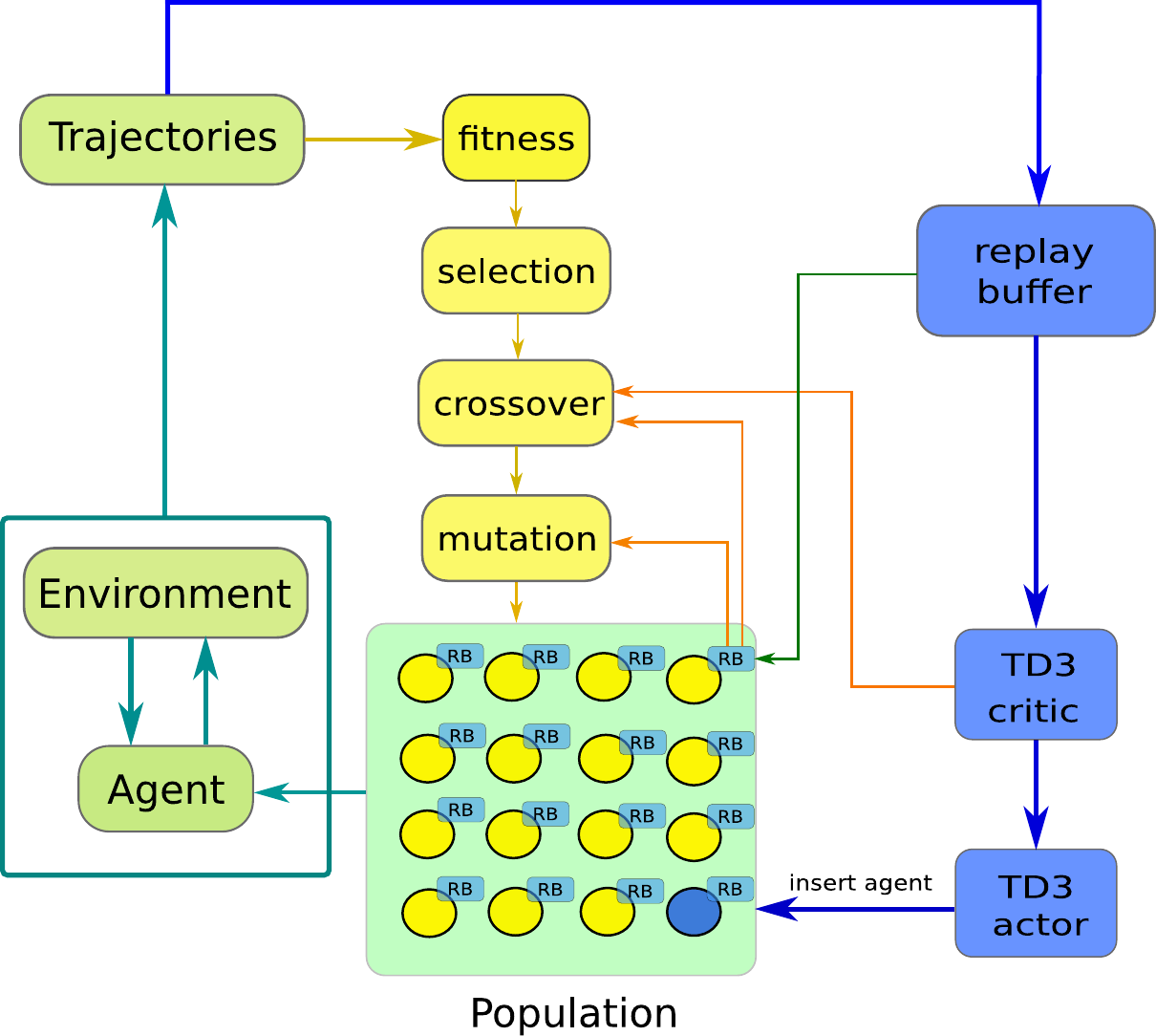}}
   \caption{
   The template architecture for \erl, \esac, \fidirl and \cerl (a) and the \pderl architecture (b). In \erl, an actor learned by \ddpg \citep{lillicrap2015continuous} is periodically injected into the population and submitted to evolutionary selection. If \ddpg performs better than the GA, this will accelerate the evolutionary process. Otherwise the \ddpg agent is just ignored. In \esac, \ddpg is replaced by \sac \citep{haarnoja2018soft} and in \fidirl, the GA is replaced by \ars \citep{mania2018simple}. In \cerl, the \ddpg agent is replaced by a set of \tddd actors sharing the same replay buffer, but each using a different discount factor. Again, those of such actors that perform better than the rest of the population are kept and enhance the evolutionary process, whereas the rest is discarded by evolutionary selection. In \pderl, the genetic operators of \erl are replaced by operators using local replay buffers so as to better leverage the step-based experience of each agent. \label{fig:erl_pderl}}
   \end{center}
\end{figure}

One of the main algorithms at the origin of the renewal of combining evolution and RL is \erl \citep{khadka2018evolutionaryNIPS}, see \figurename~\ref{fig:erl_cerl}. It was published simultaneously with the \ggac and \ggppo algorithms \citep{chang2018genetic} but its impact was much greater. Its combination mechanism consists in injecting the RL actor into the evolutionary population.

The \erl algorithm was soon followed by \cerl \citep{khadka2019collaborative} which extends \erl from RL to {\em distributed RL} where several agents learn in parallel, and all these agents are injected into the evolutionary population.
The main weakness of \erl and \cerl is their reliance on a genetic algorithm which applies a standard $n$-point-based crossover and a Gaussian weight mutation operator to a direct encoding of the neural network architecture as a simple vector of parameters. This approach is known to require tedious hyperparameter tuning and generally perform worse than evolution strategies which are also mathematically more founded \citep{salimans2017evolution}. 
In particular, the genetic operators used in \erl and \cerl based on a direct encoding have been shown to induce a risk of catastrophic forgetting of the behavior of efficient individuals. 

The \pderl algorithm \citep{bodnar2020proximal}, see \figurename~\ref{fig:pderl}, builds on this criticism and proposes two alternative evolution operators. Instead of standard crossover, all agents carry their own replay buffer and crossover selects the best experience in both parents to fill the buffer of the offspring, before applying behavioral cloning to get a new policy that behaves in accordance with the data in the buffer. This operator is inspired by the work of \cite{gangwani2017policy}. For mutation, they take as is the improved operator proposed in \cite{lehman2018safe}, which can be seen as applying a Gauss-Newton method to perform the policy gradient step \citep{pierrot2018first}.

Another follow-up of \erl is the \esac algorithm \citep{suri2020maximum}. It uses the \sac algorithm instead of \ddpg and a modified evolution strategy instead of a genetic algorithm, but the architecture follows the same template. 
Similarly, the \fidirl algorithm \citep{shi2019fidi} combines \ddpg with Augmented Random Search (\ars), a finite difference algorithm which can be seen as as simplified version of evolution strategies \citep{mania2018simple}. \fidirl uses the \erl architecture as is. The method is shown to outperform \ars alone and \ddpg alone, but neither \esac nor \fidirl are compared to any other combination listed in this survey.
Finally, the \xddpg algorithm is a version of \erl with several asynchronous \ddpg actors where the buffers from the evolutionary agents and from the \ddpg agents are separated, and the most recent \ddpg agent is injected into the evolutionary population at each time step \citep{espositi2020gradient}.

The \bnet algorithm \citep{stork2021behavior} is borderline in this survey as it does not truly use an RL algorithm, but uses a Behavior-Based Neuroevolution (BBNE) mechanism which is only loosely inspired from RL algorithms, without relying on gradient descent. \bnet combines a robust selection method based on standard fitness, a second mechanism based on the advantage of the behavior of an agent, and a third mechanism based on a surrogate estimate of the return of policies. The BBNE mechanism is reminiscent of the Advantage Weighted Regression (\awr) algorithm \citep{peng2019advantage}, but it uses an evolutionary approach to optimize this behavior-based criterion instead of standard gradient-based methods. The reasons for this choice is that the evolutionary part relies on Cartesian Genetic Programming \citep{miller2009cartesian} which evolves the structure of the neural networks, but gradient descent operators cannot be applied to networks whose structure is evolving over episodes.

\begin{figure}[!ht]
  \begin{center}
\subfloat[\label{fig:cspc}]{\includegraphics[height=0.33\linewidth]{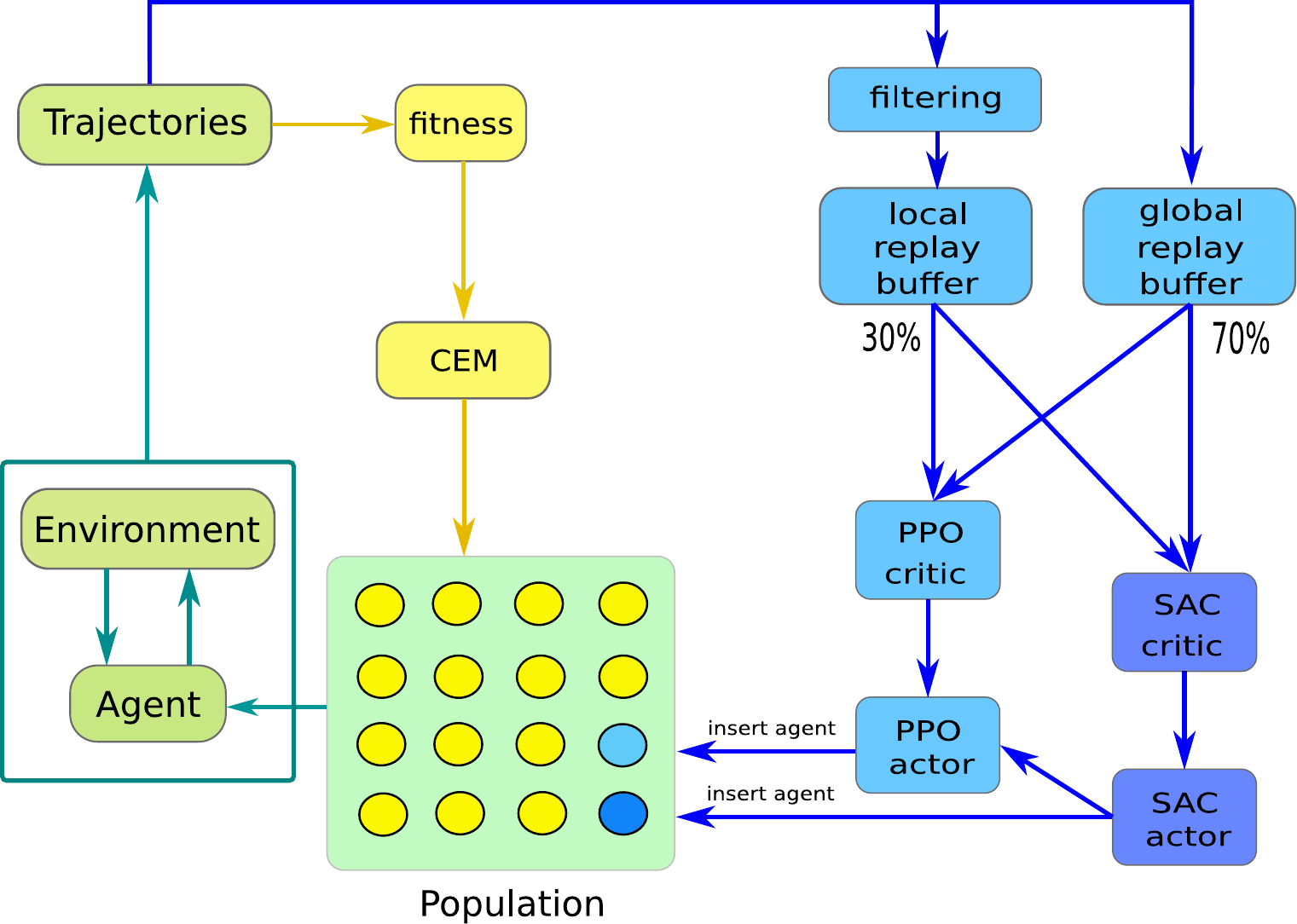}}
\hspace{0.4cm}
    \subfloat[\label{fig:cemrl}]{\includegraphics[height=0.33\linewidth]{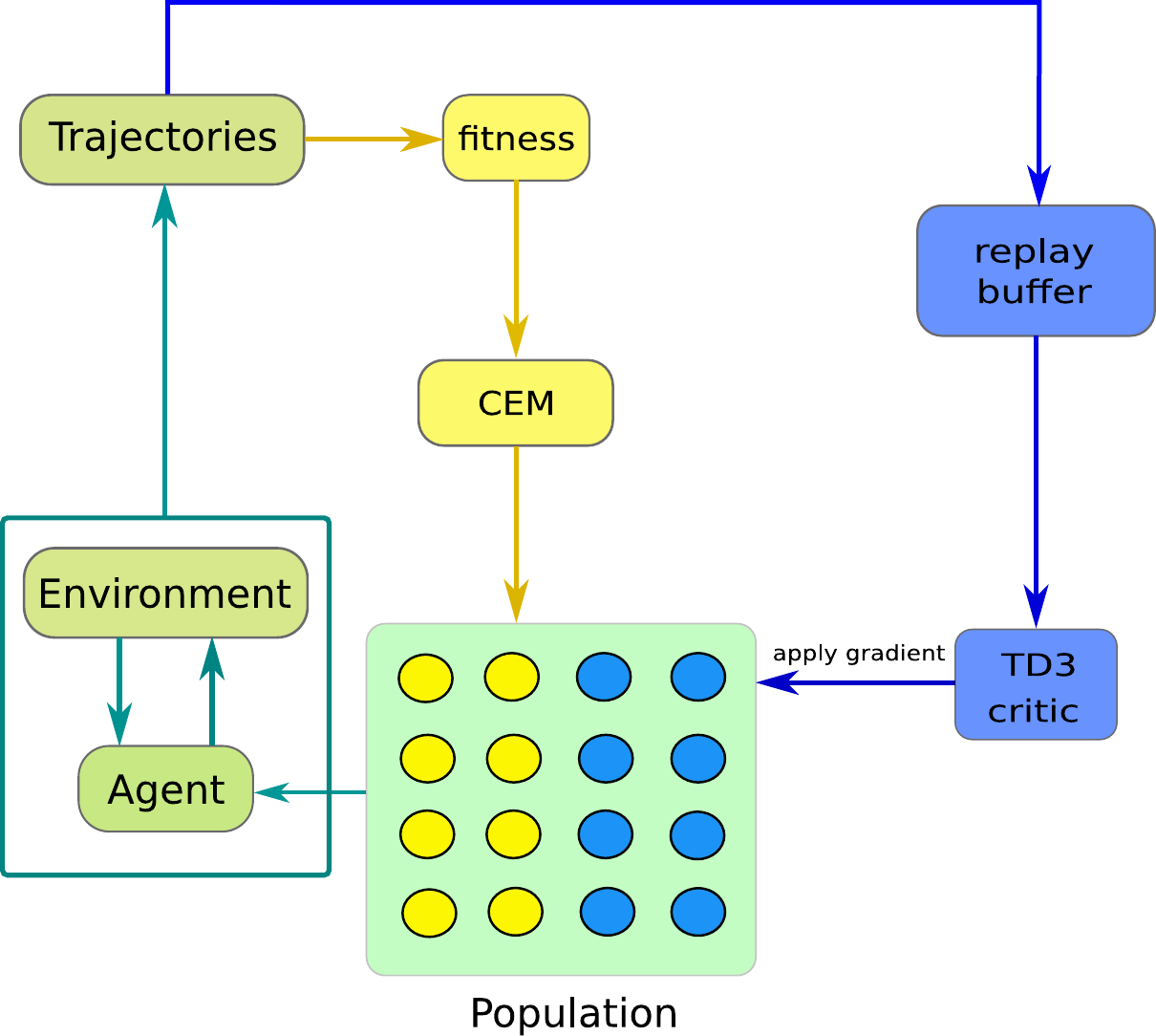}}
  \hspace{0.7cm}
   \caption{
   The \cspc (a) and \cemrl (b) architectures. In \cspc, an on-policy and an off-policy algorithms are combined, together with two replay buffers and a performance-based actor injection rule, to improve the sample efficiency of \erl-like methods. In \cemrl, gradient steps from the \tddd critic are applied to half the population of evolutionary agents. If applying this gradient is favorable, the corresponding individuals are kept, otherwise they are discarded. \label{fig:cspc_cemrl}}
   \end{center}
\end{figure}

The \chdrl architecture \citep{zheng2020cooperative} extends the \erl approach in several ways to improve the sample efficiency of the combination. First, it uses two levels of RL algorithms, one on-policy and one off-policy, to benefit from the higher sample efficiency of off-policy learning. Second, instead of injecting an actor periodically in the evolutionary population, it does so only when the actor to be injected performs substantially better than the evolutionary agents. Third, it combines the standard replay buffer with a smaller local one which is filled with filtered data to ensure using the most beneficial samples. The \cspc algorithm, depicted in \figurename~\ref{fig:cspc} is an instance of \chdrl using the \sac and \ppo \citep{schulman2017proximal} algorithms.

Note that if an RL actor is injected in an evolutionary population and if evolution uses a direct encoding, the RL actor and evolution individuals need to share a common structure. Removing this constraint might be useful, as evolutionary methods are often applied to smaller policies than RL methods. For doing so, one might call upon any policy distillation mechanism that strives to obtain from a large policy a smaller policy with similar capabilities.

\subsection{RL gradient addition}
\label{sec:rga}

Instead of injecting an RL actor into the population, another approach consists in applying gradient steps to some members of this population. This is the approach of the \cemrl algorithm \citep{pourchot2018cem}, see \figurename~\ref{fig:cemrl}. This work was followed by \cemacer \citep{tang2021guiding} which simply replaces \tddd \citep{fujimoto2018addressing} with \acer \citep{wang2016sample}.

\subsection{Evolution from the RL actor}

\begin{figure}[!ht]
  \begin{center}
    \subfloat[\label{fig:ggn}]{\includegraphics[width=0.36\linewidth]{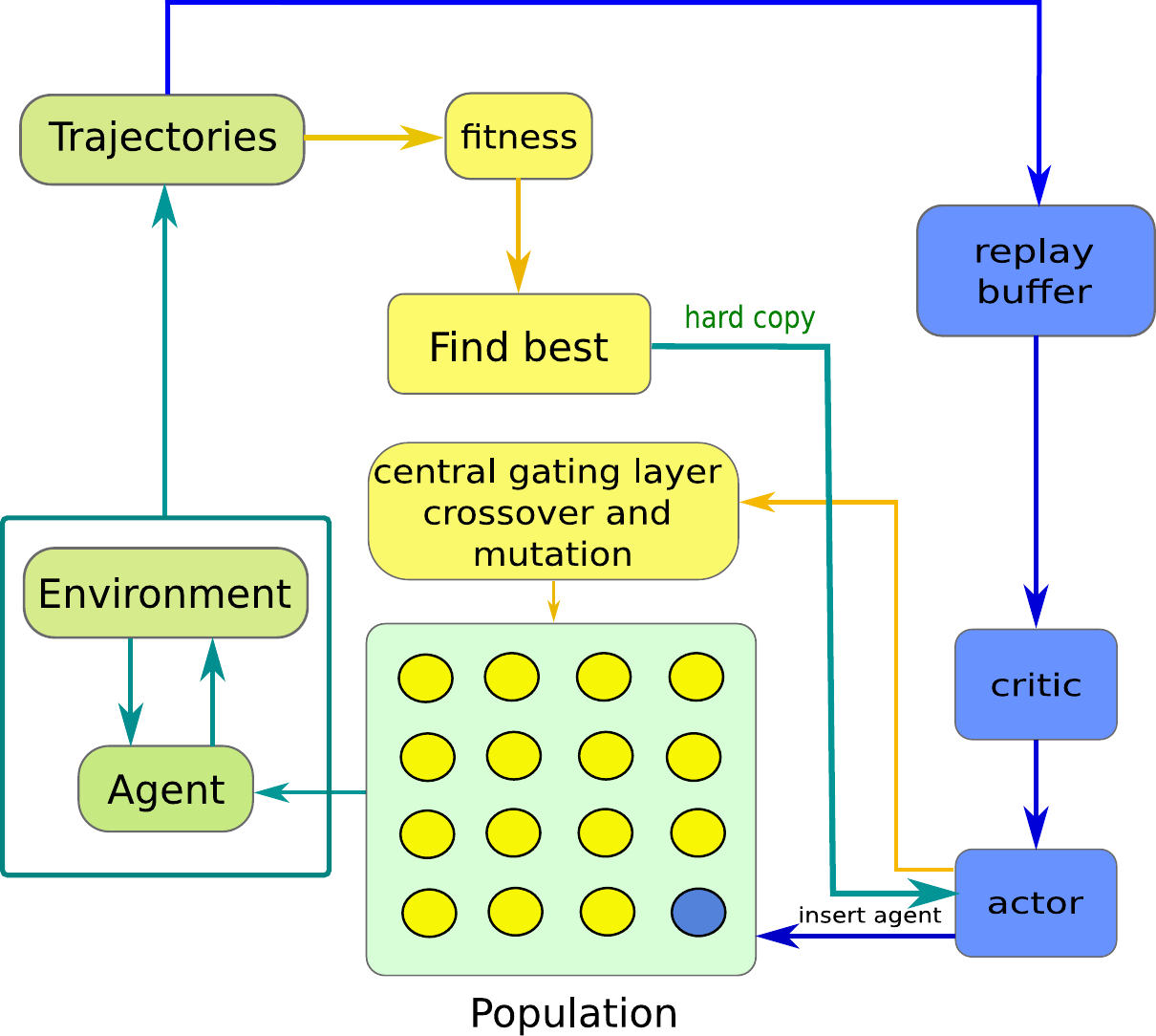}}
    \hspace{0.7cm}
    \subfloat[\label{fig:superl}]{\includegraphics[width=0.36\linewidth]{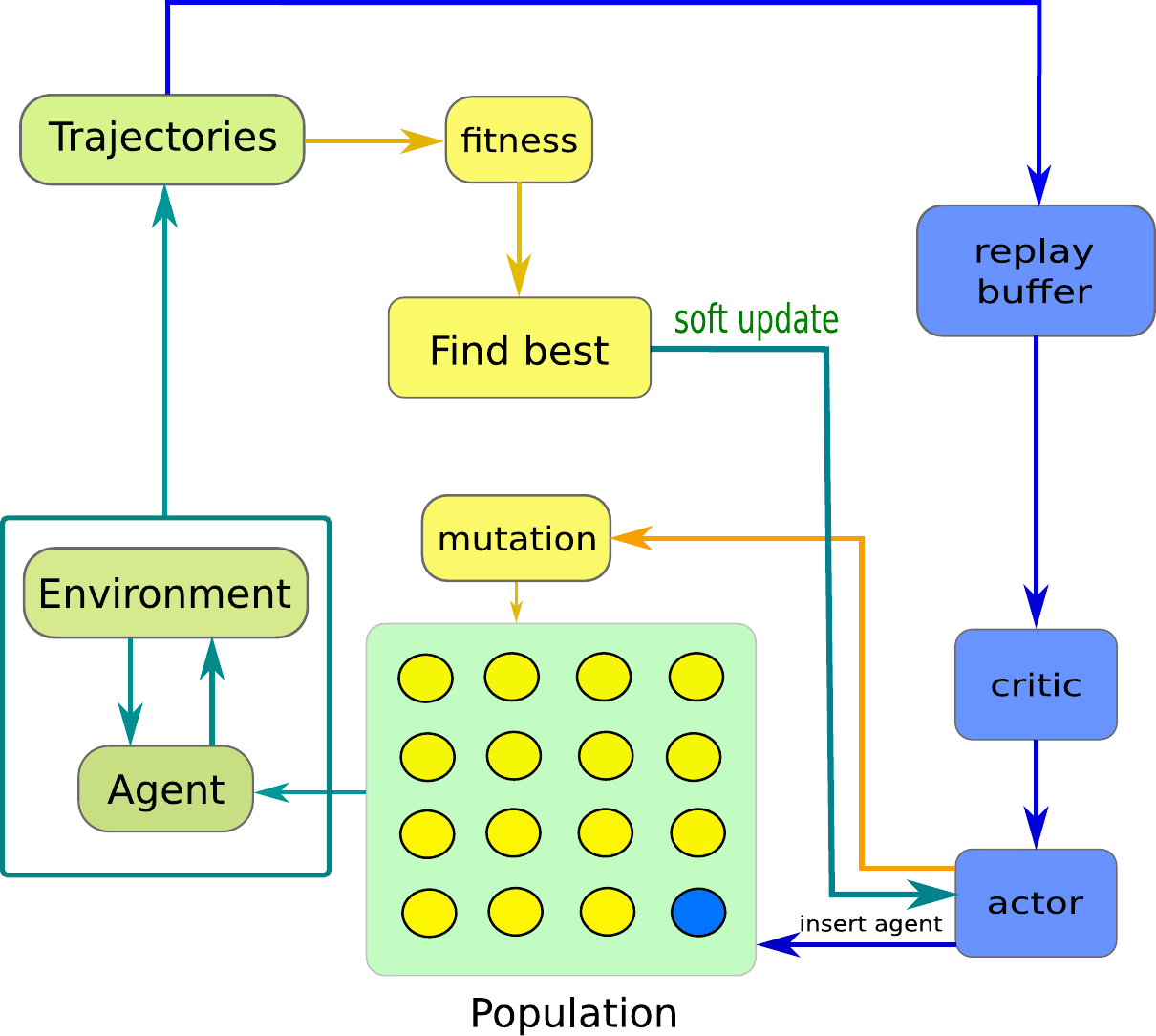}}

   \caption{
   In the \ggn (a) and \superl (b) architectures, the evolutionary population is built locally from the RL actor. In \ggn, the evolutionary part explores the structure of the central layer of the actor network. In \superl, more standard mutations are applied, the non-mutated actor is inserted in the evolutionary population and the actor is soft-updated towards its best offspring.  \label{fig:suprl_ggn}}
   \end{center}
\end{figure}

In the algorithms listed so far, the main loop is evolutionary and the RL loop is used at a slower pace to accelerate it. In the \ggn \citep{chang2018genetic} and \superl \citep{marchesini2021genetic} algorithms, by contrast, the main loop is the RL loop and evolution is used to favor exploration. 

In \ggn, shown in \figurename~\ref{fig:ggn}, evolution is used to activate or deactivate neurons of the central layer in the architecture of the actor according to a binary genome. By sampling genomes using evolutionary operators, various actor architectures are evaluated and the one that performs best benefits from a critic gradient step, before its genome is used to generate a new population of architectures. This mechanism provides a fair amount of exploration both in the actor structures and in the generated trajectories and outperforms random sampling of the genomes. Two instances of the \ggn approach are studied, \ggac based on \aac and \ggppo based on \ppo, and they both outperform the RL algorithm they use.

The \superl algorithm, shown in \figurename~\ref{fig:superl}, is similar to \ggn apart from the fact that evolving the structure of the central layer is replaced by performing standard Gaussian noise mutation of all the parameters of the actor. Besides, if one of the offspring is better than the current RL agent, the latter is modified towards this better offspring through a soft update mechanism. Finally, the non-mutated actor is also inserted in the evolutionary population, which is not the case in \ggn.

\subsection{Using a surrogate fitness}

\begin{figure}[!ht]
  \begin{center}
  \subfloat[\label{fig:scerl}]{\includegraphics[width=0.36\linewidth]{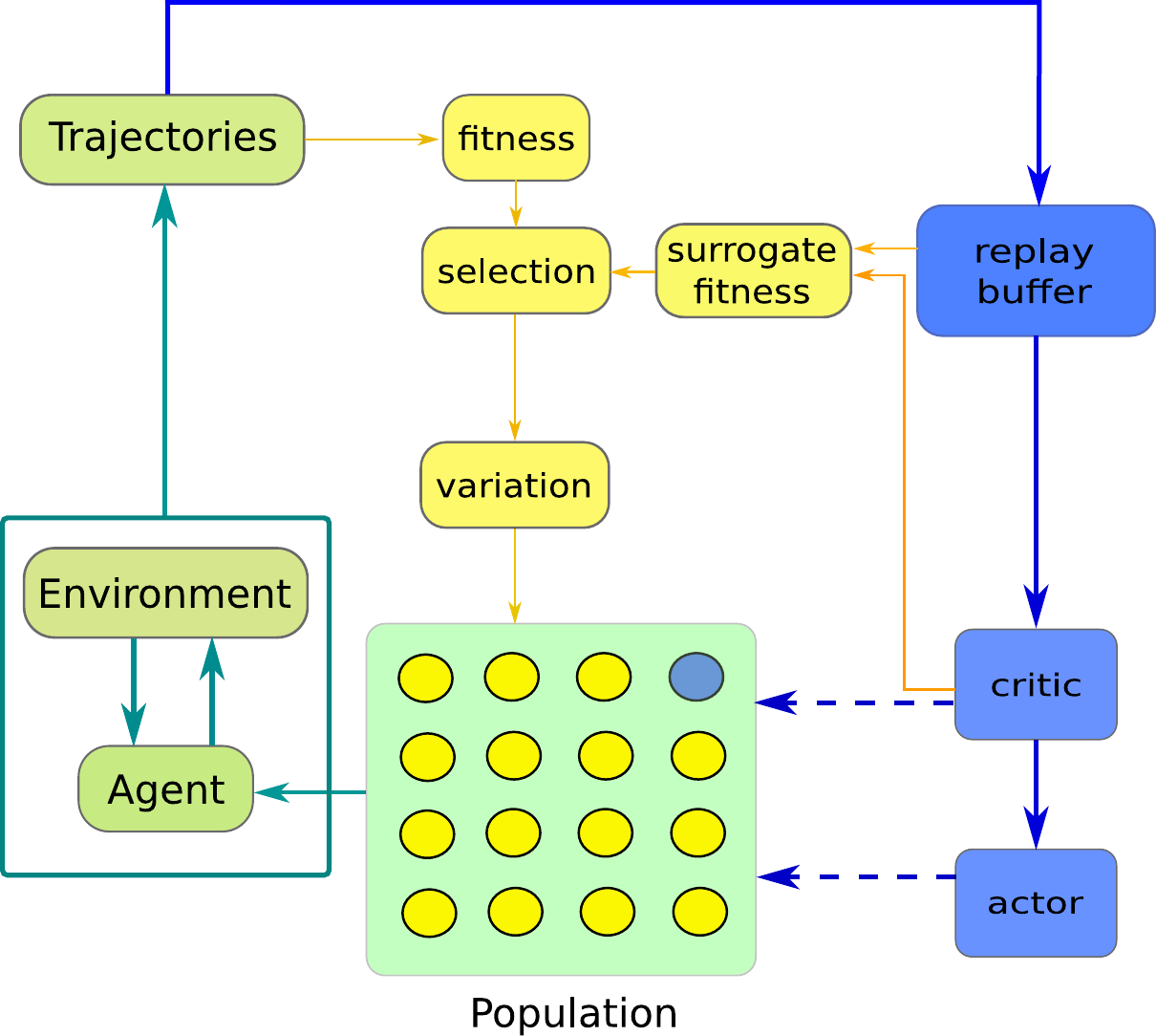}}
  \hspace{0.7cm}
  \subfloat[\label{fig:pgps}]{\includegraphics[width=0.36\linewidth]{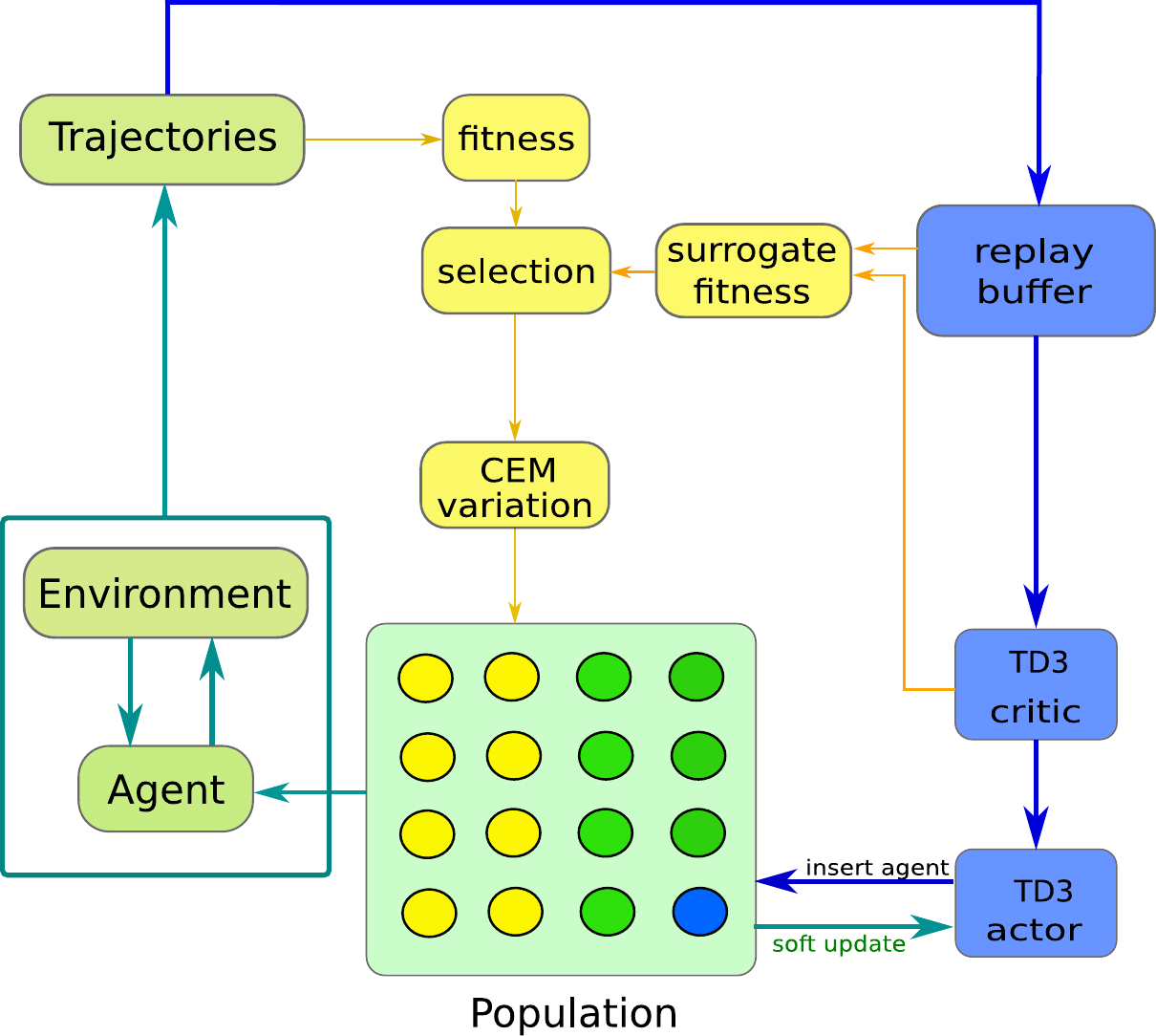}}
   \caption{
    The \scerl (a) and \pgps (b) architectures are two approaches to improve sample efficiency by using a critic network as a surrogate for evaluating evolutionary individuals. In \scerl, the surrogate control part is generic and can be applied to several architectures such as \erl, \cerl or \cemrl. It considers the critic as a surrogate model of fitness, making it possible to estimate the fitness of a new individual without generating additional samples. (b) The \pgps uses the same idea but combines it with several other mechanisms, such as performing a soft update of the actor towards the best evolutionary agent or filling half the population using the surrogate fitness and the other half from \cem generated agents. \label{fig:scerl_pgps}}
   \end{center}
\end{figure}

A weakness of all the methods combining evolution and RL that we have listed so far is that they require evaluating the agents to perform the evolutionary selection step, which may impair sample efficiency. In the \scerl \citep{wang2022surrogate} and \pgps \citep{kim2020pgps} architectures, this concern is addressed by using a critic network as a surrogate for evaluating an agent. Importantly, the evaluation of individuals must initially rely on the true fitness but can call upon the critic more and more often as its accuracy gets better. As shown in \figurename~\ref{fig:scerl}, the \scerl architecture is generic and can be applied on top of any of the combinations we have listed so far. In practice, it is applied to \erl, \pderl and \cemrl, resulting in the \serl and \spderl algorithms in the first two cases.

The \pgps algorithm \citep{kim2020pgps}, shown in \figurename~\ref{fig:pgps}, builds on the same idea but uses it in the context of a specific combination of evolutionary and RL mechanisms which borrows ideas from several of the previously described methods. In more details, half of the population is filled with agents evaluated from the surrogate fitness whereas the other half are generated with \cem. Furthermore, the current \tddd actor is injected into the population and benefits from a soft update towards the best agent in the population.

\section{Evolution of actions for performance}
\label{sec:actions}

In this section we cover algorithms where evolution is used to optimize an action in a given state, rather than optimizing policy parameters. The general idea is that variation-selection methods such as \cem can optimize any vector of parameters given some performance function of these parameters. In the methods listed in the previous section, the parameters were those of a policy and the performance was the return of that policy. In the methods listed here, the parameters specify the action in a given state and the performance is the Q-value of this action in that state.

In an RL algorithm like \ql, the agent needs to find the action with the highest value in a given state for two things: for performing critic updates, that is updating its estimates of the action-value function using $Q(s_t,a_t) \leftarrow r(s_t,a_t) + \max_a Q(s_{t+1},a) - Q(s_t,a_t)$, and for acting using $\argmax_a Q(s_t,a)$. When the action space is continuous, this amounts to solving an expensive optimization problem, and this is required at each training step.
The standard solution to this problem in actor-critic methods consists in considering the action of the actor as a good proxy for the best action. The estimated best action, that we note $\bar{a_t}$, is taken to be the actor's action $\bar{a_t} = \pi(s_t)$, resulting in using $Q(s_t,a_t) \leftarrow r(s_t,a_t) + \max_a Q(s_{t+1}, \pi(s_{t+1})) - Q(s_t,a_t)$ for the critic update and using $\bar{a_t}$ for acting.

But as an alternative, one can call upon a variation-selection method to find the best performing action over a limited set of sampled actions. This approach is used in the \qtopt algorithm \citep{kalashnikov2018qt}, as well as in the \cgp \citep{simmons2019q}, \saccepo \citep{shi2021soft}, \grac \citep{shao2021grac} and \easrl \citep{ma2022evolutionary} algorithms. This is the approach we first cover in this section. The \zospi algorithm \citep{sun2020zeroth} also benefits from optimizing actions with a variation-selection method, though it stems from a different perspective.

\begin{table*}[htp] 
\centering
\caption{Combinations evolving actions for performance. The cells in green denote where evolutionary optimization takes places. We specify the use of \cem for optimizing an action with $\bar{a_t} =\text{\cem}(\textit{source}, N, N_e, I)$, where $\textit{source}$ is the source from which we sample initial actions, $N$ is the size of this sample (the population), $N_e$ is the number of elite solutions that are retained from a generation to the next and $I$ is the number of iterations. For \pso, the shown parameters are the number of action $N$ and the number of iterations $T$. And we use $\bar{a_t}=\argmax(\textit{source}, N)$ for simply take the best action over $N$ samples from a given $\textit{source}$.} \label{tab:algos2}
\resizebox{\textwidth}{!}{\begin{tabular}{|r|l|l|l|}
    \hline
    \backslashbox{Algo.}{Prop.} & Critic update & Action Selection & Policy Update \\  \hline
    \multicolumn{1}{|l|}{\qtopt \cite{kalashnikov2018qt}} & \cellcolor{green!30}$\bar{a_t} =$\cem(random, 64, 6, 2) & \cellcolor{green!30}$\bar{a_t} =$\cem(random, 64, 6, 2) & No policy \\ \hline
    \multicolumn{1}{|l|}{\cgp \cite{simmons2019q}} & \cellcolor{green!30}$\bar{a_t} =$\cem(random, 64, 6, 2) & $\bar{a_t} = \pi(s_t)$ & BC or DPG \\ \hline
    \multicolumn{1}{|l|}{\easrl \cite{ma2022evolutionary}} & \cellcolor{green!30}$\bar{a_t} =$\pso(10,10) & $\bar{a_t} = \pi(s_t)$ & BC + DPG   \\ \hline
    \multicolumn{1}{|l|}{\saccepo \cite{shi2021soft}} & \sac update & \cellcolor{green!30}$\bar{a_t} =$\cem($\pi$, 60 \shortrightarrow 140, $3\%$ \shortrightarrow $7\%$, 6 \shortrightarrow 14) & BC \\ \hline
    \multicolumn{1}{|l|}{\grac \cite{shao2021grac}} & \cellcolor{green!30}$\bar{a_t} =$\cem($\pi$, 256, 5, 2) & \cellcolor{green!30}$\bar{a_t} =$\cem($\pi$, 256, 5, 2) & PG with two losses \\ \hline
    \multicolumn{1}{|l|}{\zospi \cite{sun2020zeroth}} &  \ddpg update & $\bar{a_t} = \pi(s_t)$ + perturb. network &\cellcolor{green!30}BC($\bar{a_t}=\argmax(random,50)$) \\ \hline
\end{tabular}}
\end{table*}

As Table~\ref{tab:algos2} shows, the \qtopt algorithm \citep{kalashnikov2018qt} {\bf simply samples 64 random actions} in the action space and performs two iterations of \cem to get a high performing action, both for critic updates and action selection. It is striking that such a simple method can perform well even in large action spaces.
This simple idea was then improved in the \cgp algorithm \citep{simmons2019q} so as to {\bf avoid the computational cost of action inference}. Instead of using \cem to sample an action at each time step, a policy network is learned based on the behavior of the \cem. This network can be seen as a surrogate of the \cem sampling process and is trained either from the sampled $\bar{a_t}$ using Behavioral Cloning (BC) or following a Deterministic Policy Gradient (DPG) step from the critic.

The \easrl algorithm \citep{ma2022evolutionary} is similar to \cgp apart from the fact that it uses Particle Swarm Optimization (\pso) instead of \cem.
Besides, depending on the sign of the advantage of the obtained action $\bar{a_t}$, it uses either BC or DPG to update the policy for each sample.

Symmetrically to \cgp, the \saccepo algorithm \citep{shi2021soft} performs standard critic updates using \sac but selects actions using \cem. More precisely, it {\bf introduces the idea to sample the action from the current policy rather than randomly}, and updates this policy using BC from the sampled actions. Besides, the paper investigates the effect of the \cem parameters but does not provide solid conclusions.

The \grac algorithm \citep{shao2021grac} {\bf combines ideas from \cgp and \saccepo}. A stochastic policy network outputs an initial Gaussian distribution for the action at each step. Then, a step of \cem drawing 256 actions out of this  distribution is used to further optimize the choice of action both for critic updates and action selection. The policy itself is updated with a combination of two training losses.

Finally, the \zospi algorithm \citep{sun2020zeroth} {\bf calls upon variation-selection for updating the policy} rather than for updating the critic or selecting the action. Its point is rather than gradient descent algorithms tend to get stuck into local minima and may miss the appropriate direction due to various approximations, whereas a variation-selection method is more robust. Thus, to update its main policy, \zospi simply samples a set of actions and performs BC towards the best of these actions, which can be seen as a trivial variation-selection method. The typical number of sampled actions is 50. It then adds a policy perturbation network to perform exploration, which is trained using gradient descent.

\section{Evolution of policies for diversity}
\label{sec:diversity}

\begin{figure}[!ht]
  \begin{center}
\subfloat[\label{fig:ns_rl}]{\includegraphics[width=0.36\linewidth]{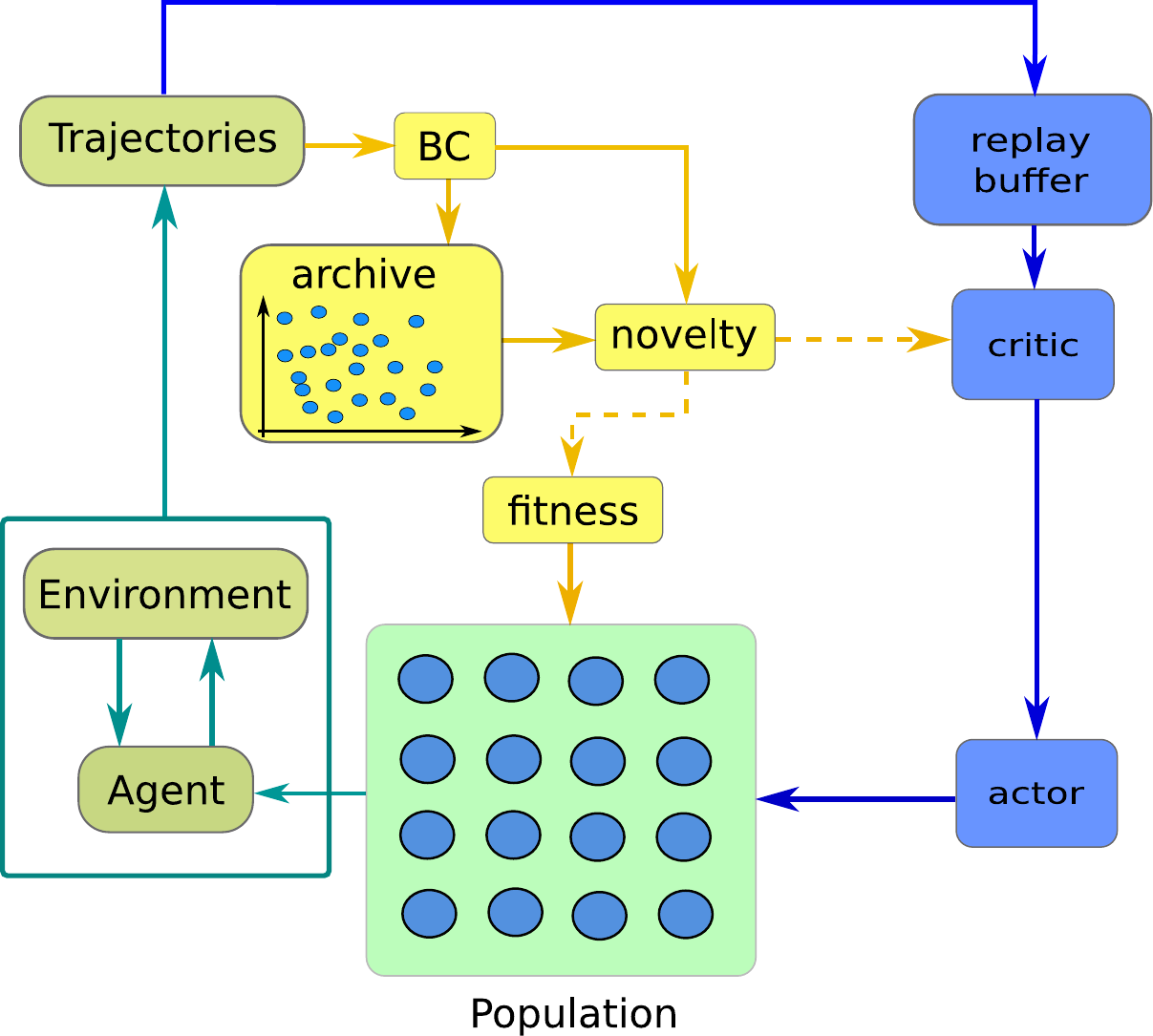}}
     \hspace{0.7cm}
\subfloat[\label{fig:me}]{\includegraphics[width=0.36\linewidth]{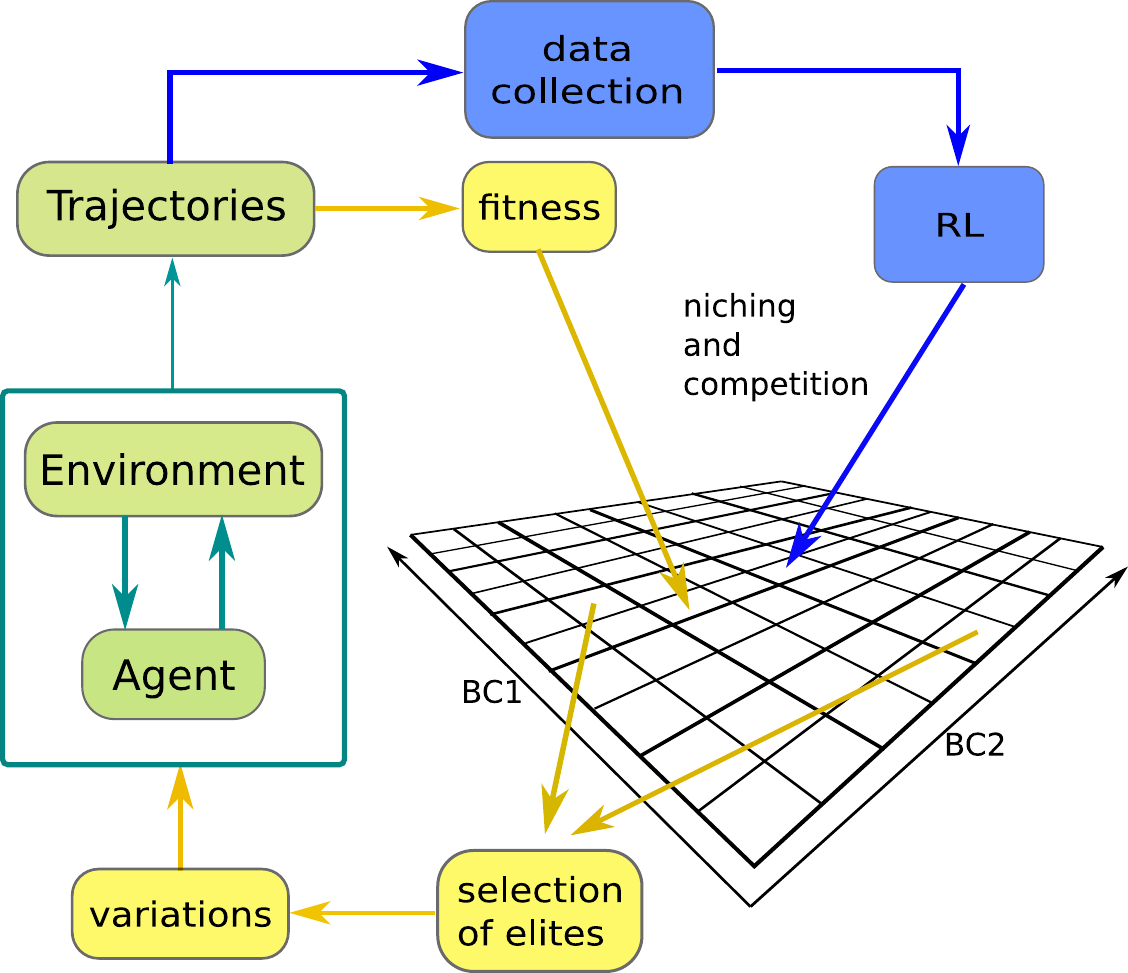}} 
   \caption{Template architectures for combining deep RL with novelty search (a) and quality-diversity (b). The latter builds on Fig.~2 in \cite{mouret2020evolving}. Both architectures rely on a behavioral characterization space and maintain an archive in that space. \label{fig:ns_qd}}
   \end{center}
\end{figure}

The trade-off between exploration and exploitation is central to RL. In particular, when the reward signal is sparse, efficient exploration becomes crucial. All the papers studied in this survey manage a population of agents, hence their capability to explore can benefit from maintaining behavioral diversity between the agents. 
This idea of maintaining behavioral diversity is central to two families of diversity seeking algorithms, the novelty search (NS) \citep{lehman2011abandoning} algorithms which do not use the reward signal at all, see \figurename~\ref{fig:ns_rl}, and the quality-diversity (QD) algorithms \citep{pugh2016quality, cully2017quality}, see \figurename~\ref{fig:me}, which try to maximize both diversity and performance. 
As the NS approach only looks for diversity, it is better in the absence of reward, or when the reward signal is very sparse or deceptive as the best one can do in the absence of reward is try to cover a relevant space as uniformly as possible \citep{doncieux2019novelty}. By contrast, the QD approach is more appropriate when the reward signal can contribute to the policy search process.
In this section we cover both families separately.

\subsection{Novelty seeking approaches}


\begin{table*}[htp]
\centering
\caption{Combinations evolving policies for diversity.  NS: Novelty Search. Policy params: distance is computed in the policy parameters space. GC-\ddpg: goal-conditioned \ddpg. Manual BC: distances are computed in a manually defined behavior characterization space.\label{tab:algos3}}
\begin{tabular}{|r|r|r|r|}
    \hline
    \backslashbox{Algo.}{Prop.} &\makecell{RL\\ algo.} & \makecell{Diversity\\ algo.} & \makecell{Distance\\ space} \\  \hline
    \multicolumn{1}{|l|}{\pssstddd \cite{jung2020population}} & \tddd & Find best & Policy params.  \\ \hline
    \multicolumn{1}{|l|}{\deprl \cite{liu2021diversity}} & \tddd & \cem & Policy params.  \\ \hline
    \multicolumn{1}{|l|}{\arac \cite{doan2019attraction}} & \sac & NS-like & Policy params.  \\ \hline
    \multicolumn{1}{|l|}{\nsrl \cite{shi2020efficient}} & GC-\ddpg & True NS &Manual BC  \\ \hline
    \multicolumn{1}{|l|}{\pnsrl \cite{liu2018pns}} & \tddd & True NS & Manual BC  \\ \hline\end{tabular}
\end{table*}

\begin{figure}[!ht]
  \begin{center}
\subfloat[\label{fig:svpg}]{\includegraphics[width=0.36\linewidth]{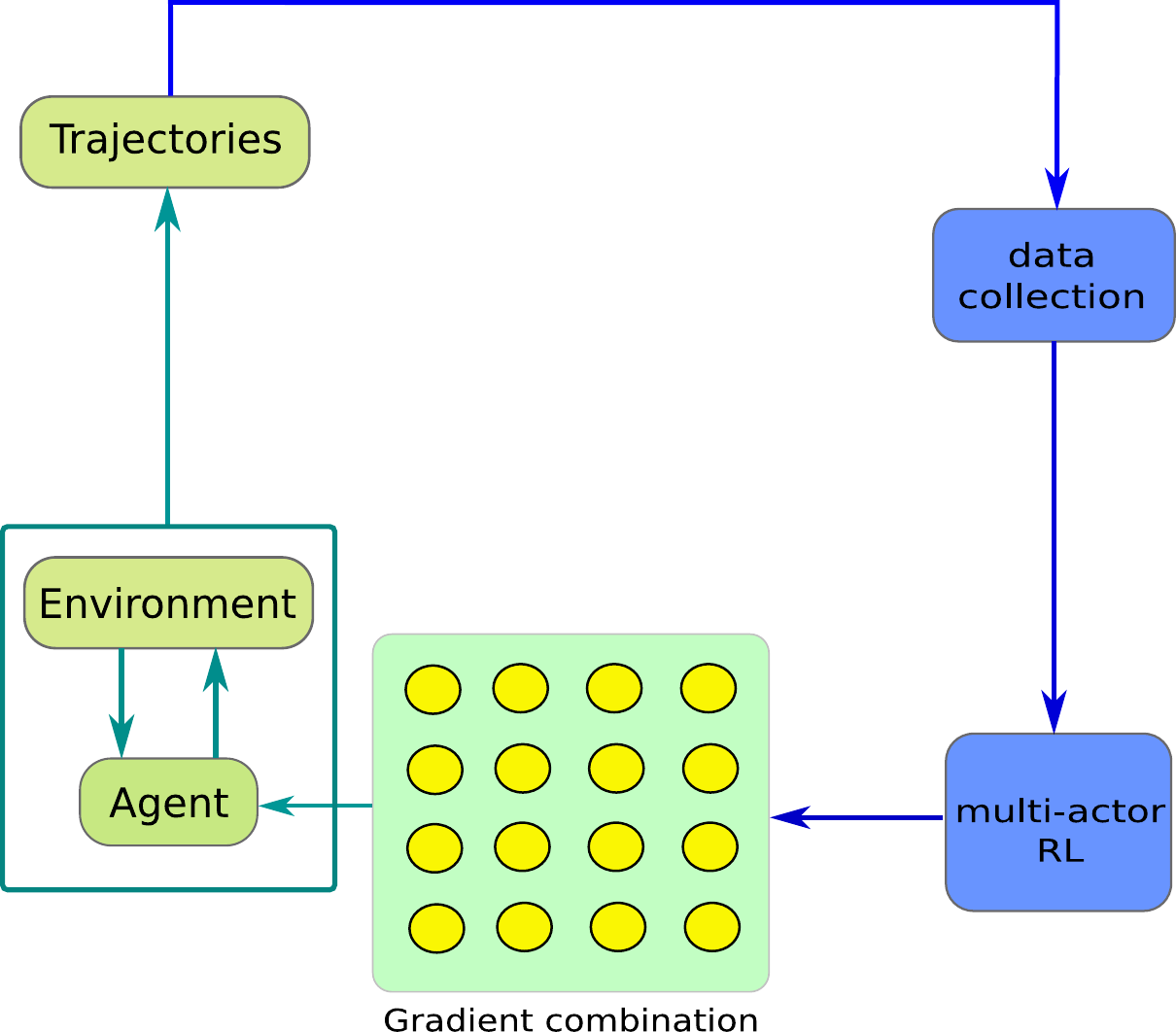}}
  \hspace{0.7cm}
\subfloat[\label{fig:dvd}]{\includegraphics[width=0.25\linewidth]{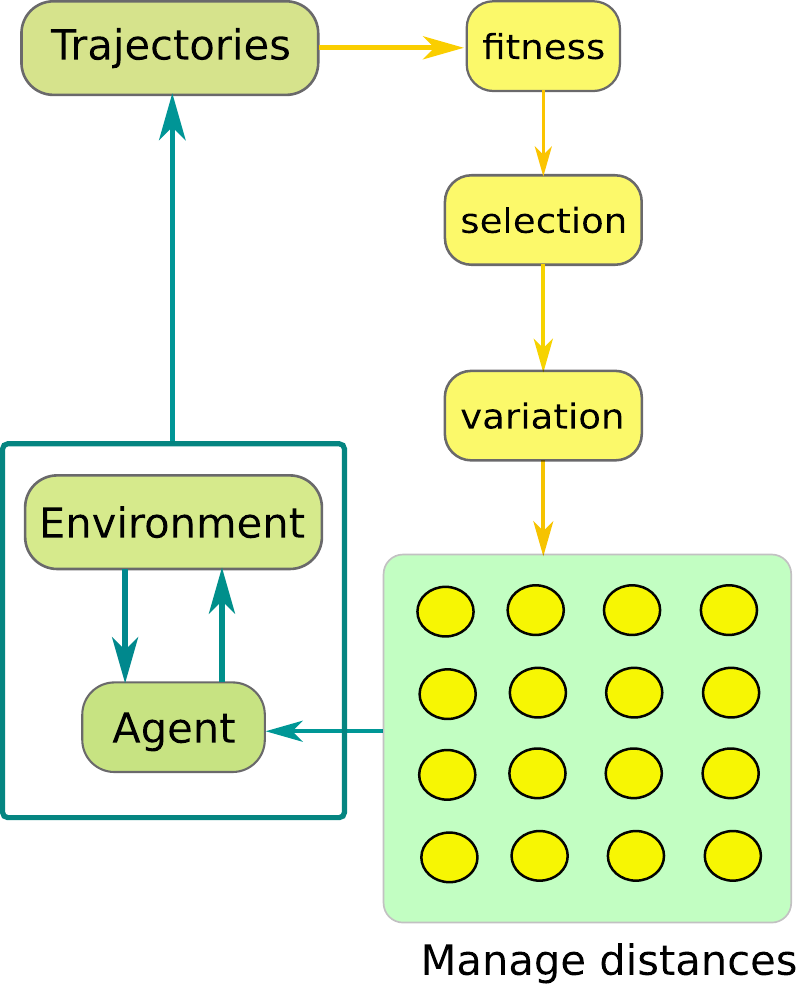}} 
  \caption{The \svpg (a) and \dvd (b) architectures.
   In \svpg, individual policy gradients computed by each agent are combined so as to ensure both diversity between agents and performance improvement.
In \dvd, a purely evolutionary approach is combined with a diversity mechanism which seeks to maximize the volume between the behavioral characterization of agents in an action embedding space. Both architectures fail to combine evolution and RL, though they both try to maximize diversity and performance in a population of agents.\label{fig:svpg_dvd}}
   \end{center}
\end{figure}

Maintaining a distance between agents in a population can be achieved in different spaces. For instance, the \svpg algorithm \citep{liu2017stein} defines distances in a kernel space and adds to the policy gradient a loss term dedicated to increasing the pairwise distance between agents. Alternatively, the \dvd algorithm \citep{parker2020effective} defines distances in an action embedding space, corresponding to the actions specified by each agent in a large enough set of random states. Then \dvd optimizes a global distance between all policies by maximizing the volume of the space between them through the computation of a determinant.
Despite their interest, these two methods depicted in \figurename~\ref{fig:svpg_dvd} do not appear in Table~\ref{tab:algos3} as the former does not have an evolutionary component and the latter uses \nsres \citep{conti2018improving} but does not have an RL component.

\begin{figure}[!ht]
  \begin{center}
\subfloat[\label{fig:p3s}]{\includegraphics[width=0.36\linewidth]{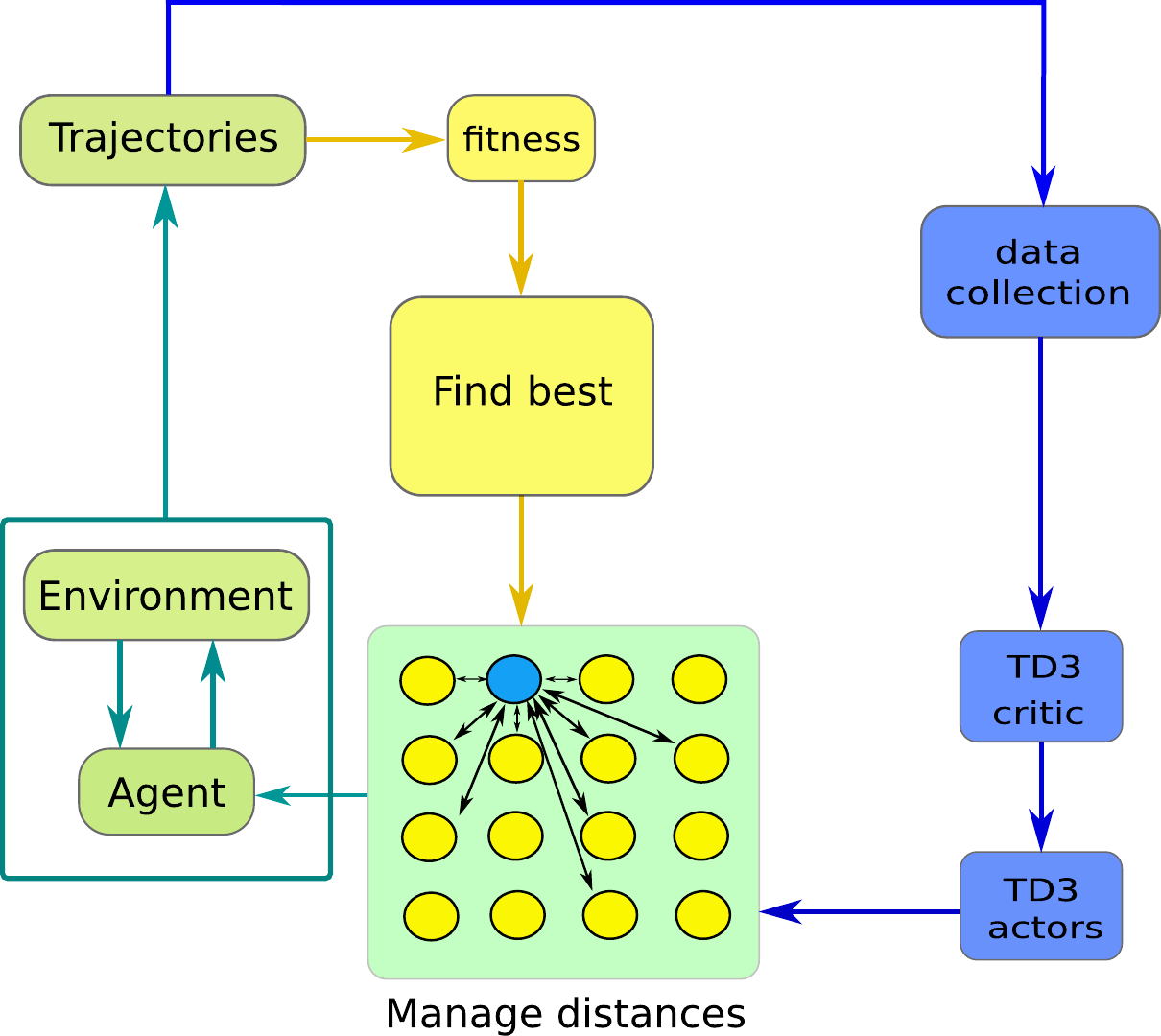}}
  \hspace{0.7cm}
\subfloat[\label{fig:arac}]{\includegraphics[width=0.36\linewidth]{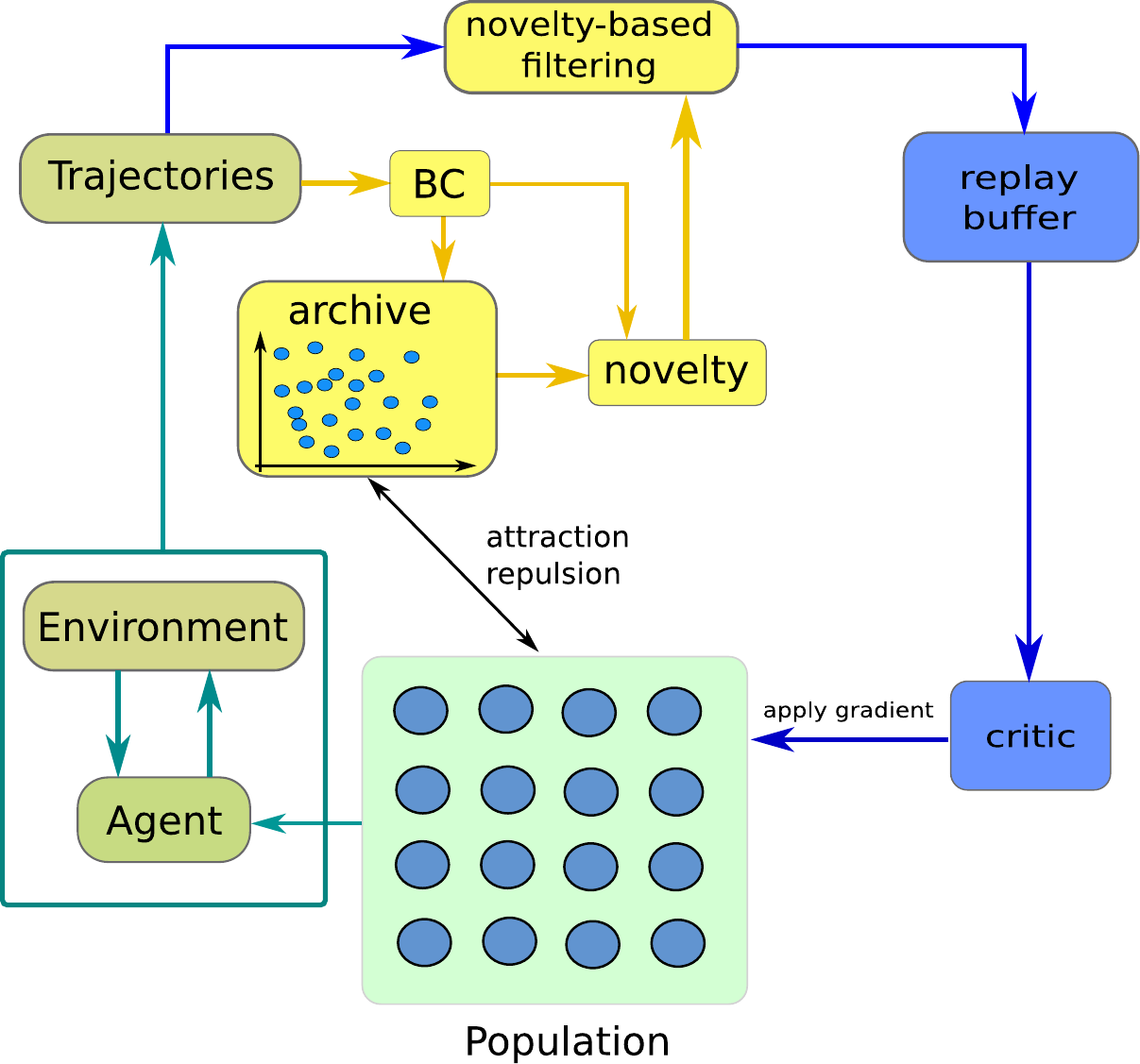}} 
   \caption{The \pssstddd (a) and \arac (b) architectures.
   In \pssstddd, all agents are trained with RL and evaluated, then they all perform a soft update towards the best agent. The \arac algorithm maintains a population of policies following a gradient from a common \sac critic \citep{haarnoja2018soft}. The critic itself is trained from trajectories of the most novel agents. Besides, diversity in the population is ensured by adding an attraction-repulsion loss $\mathcal{L}_{AR}$ to the update of the agents. This loss is computed with respect to an archive of previous agents themselves selected using a novelty criterion.\label{fig:p3s_arac}}
   \end{center}
\end{figure}

A more borderline case with respect to the focus of this survey is the \pssstddd algorithm \citep{jung2020population}. Though \pssstddd is used as a baseline in several of the papers mentioned in this survey, its equivalent of the evolutionary loop is limited to finding the best agent in the population, as shown in \figurename~\ref{fig:p3s}. This implies evaluating all these agents, but not using neither variation nor selection. Besides, the mechanism to maintain a distance between solutions in \pssstddd is ad hoc and acts in the space of policy parameters. This is also the case in the \deprl algorithm \citep{liu2021diversity}, which is just a variation of \cemrl where an ad hoc mechanism is added to enforce some distance between members of the evolutionary population.

The \arac algorithm \citep{doan2019attraction} also uses a distance in the policy parameter space, but it truly qualifies as a combination of evolution and deep RL, see \figurename~\ref{fig:arac}. 
An original feature of \arac is that it selects the data coming into the replay buffer based on the novelty of agents, which can result in saving a lot of poorly informative gradient computations. A similar idea is also present in \citep{chen2019merging} where instead of filtering based on novelty, the algorithm uses an elite replay buffer containing only the top trajectories, similarly to what we have already seen in the \cspc algorithm \citep{zheng2020cooperative}.

The methods listed so far were neither using a manually defined behavior characterization space for computing distances between agents nor an archive of previous agents to evaluate novelty. Thus they do not truly qualify as NS approaches.
We now turn to algorithms which combine both features.

Figure~\ref{fig:ns_rl} suggests that, when combining evolution and RL, novelty can be used as a fitness function, as a reward signal to learn a critic, or both.
Actually, in the two algorithms described below, it is used for both. More precisely, the RL part is used to move the rest of the population towards the most novel agent.

In the \pnsrl algorithm \citep{liu2018pns}, a group of agents is following a leader combining a standard policy gradient update and a soft update towards the leader. Then, for any agent in the group, if its performance is high enough with respect to the mean performance in an archive, it is added to the archive. Crucially, {\bf the leader is selected as the one that maximizes novelty} in the archive given a manually defined behavioral characterization. In addition, for efficient parallelization, the algorithm considers several groups instead of one, but where all groups share the same leader.

The \nsrl algorithm \citep{shi2020efficient} can be seen as a version of \cemrl whose {\bf RL part targets higher novelty} by training less novel agents to minimize in each step the distance to the BC of the most novel agent. As the most novel agent and its BC change in each iteration, {\bf the RL part is implemented with goal-conditioned policies}. This implies that the goal space is identical to the behavioral characterization space. 


\subsection{Quality-diversity approaches}
    
\begin{table*}[htp]
\centering
\caption{Quality-Diversity algorithms including an RL component. All these algorithms rely on the Map-Elites approach and the BC space is defined manually. For each algorithm in the rows, the table states whether quality and diversity are optimized using an RL approach or an evolutionary approach. \label{tab:algos3b}}
\resizebox{\textwidth}{!}{
\begin{tabular}{|r|r|r|r|r|r|}
  \hline
    \backslashbox{Algo.}{Prop.} & \makecell{Type of Archive} & \makecell{Q. improvement} & \makecell{D. improvement} \\  \hline

    \hline
    \multicolumn{1}{|l|}{\pgame \cite{nilsson2021policy}} & Map-Elites & \tddd or GA & \tddd or GA \\ \hline
        \multicolumn{1}{|l|}{\qdpg-PF \cite{cideron2020qd}} & Pareto front & \tddd  & \tddd \\ \hline
    \multicolumn{1}{|l|}{\qdpg-ME \cite{pierrot2020sample}} & Map-Elites & \tddd  & \tddd \\ \hline
    \multicolumn{1}{|l|}{\cmamega-ES \cite{tjanaka2022approximating}} & Map-Elites & \cmaes & \cmaes \\ \hline
    \multicolumn{1}{|l|}{\cmamega-(\tddd, ES) \cite{tjanaka2022approximating}} & Map-Elites & \tddd + \cmaes & \cmaes \\ \hline
\end{tabular}
}
\end{table*}

By contrast with NS approaches which only try to optimize diversity in the population, QD approaches combine this first objective with optimize the performance of registered policies, their {\em quality}. As \figurename~\ref{fig:me} suggests, when combined with an RL loop, the QD loop can give rise to various solutions depending on whether quality and diversity are improved with a evolutionary algorithm or a deep RL algorithm.

The space of resulting possibilities is covered in Table~\ref{tab:algos3b}. In more details, the \pgame algorithm \citep{nilsson2021policy} uses two optimization mechanisms, \tddd and a GA, to generate new solutions that are added to the archive if they are either novel enough or more efficient than previously registered ones with the same behavioral characterization.
By contrast, in the \qdrl approach, the mechanisms to improve quality and diversity are explicitly separated
and consist in improving a quality critic and a diversity critic using \tddd. Two implementations exist. First, the \qdpg-PF algorithm \citep{cideron2020qd} maintains a Pareto front of high quality and diversity solutions. From its side, the \qdpg-ME algorithm \citep{pierrot2020sample} maintains a Map-Elites archive and introduces an additional notion of state descriptor to justify learning a state-based quality critic.
Finally, the \cmamega approach \citep{tjanaka2022approximating} uses an ES to improve diversity and either an ES or a combination of ES and RL to improve quality. 

To summarize, one can see that both quality and diversity can be improved through RL, evolution, or both.

\section{Evolution of something else}
\label{sec:other}

In all the architecture we have surveyed so far, the evolutionary part was used to optimize either policy parameters or a set of rewarding actions in a given state. In this section, we briefly cover combinations of evolution and deep RL where evolution is used to optimize something else that matters in the RL process, or where RL mechanisms are used to improve evolution without calling upon a full RL algorithm. 
We dedicate a separate part to optimizing hyperparameters, as it is an important and active domain.

\begin{table*}[htp]
\centering
\caption{Algorithms where evolution is applied to something else than action or policy parameters, or to more than policy parameters. All algorithms in the first half optimize hyperparameters. *: The algorithm in \cite{park2021computational} is given no name. BT stands for Behavior Tree. } \label{tab:algos4}
\resizebox{\textwidth}{!}{\begin{tabular}{|r|l|l|l|}
    \hline
    \backslashbox{Algo.}{Prop.} & \makecell{RL algo.} & \makecell{Evo algo.} &\makecell{Evolves what?}  \\ 
    \hline
    \multicolumn{1}{|l|}{\gadrl \cite{sehgal2019deep, sehgal2022ga}} & \ddpg (+\her) & GA & Hyper-parameters \\ \hline
    \multicolumn{1}{|l|}{\pbt \cite{jaderberg2017population}} & Any & Ad hoc & Parameters and Hyper-parameters \\ \hline
    \multicolumn{1}{|l|}{\aacc \cite{grigsby2021towards}} & \sac & Ad hoc & Parameters and Hyper-parameters \\ \hline
    \multicolumn{1}{|l|}{\searl \cite{franke2020sample}} & \tddd & GA & Architecture, Parameters and Hyper-parameters \\ \hline
    \multicolumn{1}{|l|}{\ohtes \cite{tang2020online}} & Any & ES & Hyper-parameters \\ \hline
    \hline
    \multicolumn{1}{|l|}{\epg \cite{houthooft2018evolved}} & Ad hoc ($\sim$ \ppo) & ES & Reward-related functions \\ \hline
    \multicolumn{1}{|l|}{\eQ \cite{leite2020reinforcement}} & $\sim$ \ddpg & $\sim$ \cem & Critic  \\ \hline
    \multicolumn{1}{|l|}{\evorl \cite{hallawa2017instinct}} & \ql, \dqn, \ppo & BT & Partial policies \\ \hline
    \multicolumn{1}{|l|}{\derl \cite{gupta2021embodied}} & \ppo & GA & System's morphology \\ \hline
    \multicolumn{1}{|l|}{* \cite{park2021computational}} & \ppo & GA & System's morphology \\ \hline
\end{tabular}}
\end{table*}

\subsection{Evolution in MBRL}
\label{sec:mbrl}

The \cem algorithm can be used to optimize open-loop controllers to perform Model Predictive Control (MPC) on robotic systems in the \planet \citep{hafner2019learning} and \poplin \citep{wang2019exploring} algorithms, and an improved version of \cem for this specific context is proposed in \citep{pinneri2020sample, pinneri2020extracting}.
Besides, this approach combining open-loop controllers and MPC is seen in the \pets algorithm \cite{chua2018deep} as implementing a form of Model-Based Reinforcement Learning (MBRL), and \cem is used in \pets to choose the points from where to start MPC, improving over random shooting. Finally, in \cite{bharadhwaj2020model}, the authors propose to interleave \cem iterations and Stochastic Gradient Descent (SGD) iterations to improve the efficiency of optimization of {\bf MPC plans}, in a way reminiscent to \cemrl combining policy gradient steps and \cem steps. But all these methods are applied to an open-loop control context where true reinforcement learning algorithms can not be applied, hence they do not appear in Table~\ref{tab:algos4}. 

\subsection{Evolution of hyper-parameters}
\label{sec:pbt}

Hyperparameter optimization (HPO) is notoriously hard and often critical in deep RL. The most straightforward way to leverage evolutionary methods in this context consists in nesting the deep RL algorithm within an evolutionary loop which tunes the hyper-parameters. This is the approach of the \gadrl algorithm \citep{sehgal2019deep, sehgal2022ga}, but this obviously suffers from a very high computational cost. Note that the authors write that \gadrl uses \ddpg + \her, but the use of \her is in no way clear as the algorithm does not seem to use goal-conditioned policies.

\begin{figure}[!ht]
  \begin{center}
    \includegraphics[width=0.36\linewidth]{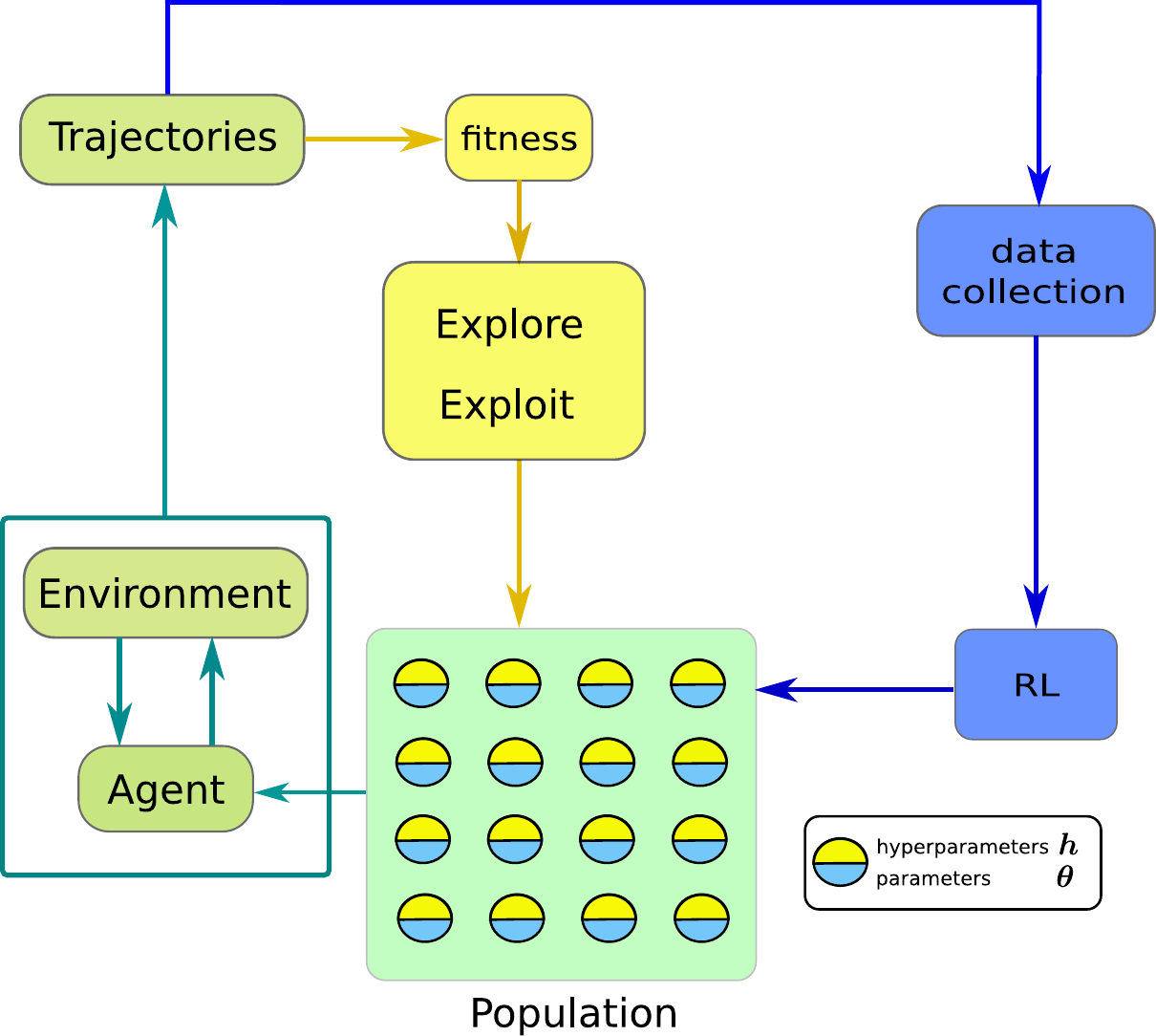}
   \caption{
   The \pbt architecture. The evolution part consists of two operators, {\em explore} and {\em exploit} which act both on the hyperparameters and the parameters of the agents. \label{fig:pbt}}
   \end{center}
\end{figure}

More interestingly, the \pbt architecture \citep{jaderberg2017population} is designed to solve this problem by combining distributed RL with an evolutionary mechanism which acts both on the parameters and hyperparameters within the RL training loop. It was successfully used in several challenging applications \citep{jaderberg2019human} and benefits from an interesting capability to {\bf tune the hyperparameters according to the current training dynamics}, which is an important meta-learning capability \citep{khamassi2017active}. A follow-up of the \pbt algorithm is the \aacc algorithm \citep{grigsby2021towards}, which basically applies the same approach but with a better set of hyperparameters building on lessons learned in the recent deep RL literature.

A limitation of \pbt is that each actor uses its own replay buffer. Instead, in the \searl algorithm \citep{franke2020sample}, {\bf the experience of all agents is shared into a unique buffer}. Furthermore, \searl~ {\bf simultaneously performs HPO and Neural Architecture Search}, resulting in better performance than \pbt.
Finally, the \ohtes algorithm \citep{tang2020online} also uses a shared replay buffer, but limits the role of evolution to optimizing hyperparameters and does so with an ES algorithm. Given the importance of the problem, there are many other HPO methods, most of which are not explicitly calling upon an evolutionary approach. For a wider survey of the topic, we refer the reader to \cite{parker2022automated}.



\subsection{Evolution of miscellaneous RL or control components}
\label{sec:rest}

Finally, we briefly survey the rest of algorithms listed in Table~\ref{tab:algos4}. The \epg algorithm \citep{houthooft2018evolved} uses a meta-learning approach to evolve {\bf the parameters of a loss function} that replaces the policy gradient surrogate loss in policy gradient algorithms. The goal is to find a reward function that will maximize the capability of an RL algorithm to achieve a given task. A consequence of its design is that it cannot be applied to an actor-critic approach.

Instead of evolving a population of agents, the \eQ algorithm \citep{leite2020reinforcement} evolves {\bf a population of critics}, which are fixed over the course of learning for a given agent. This is somewhat symmetric to the previously mentioned \zoac algorithm \citep{lei2022zeroth} which uses evolution to update an actor given a critic trained with RL.

The \evorl algorithm \citep{hallawa2021evo} evolves {\bf partial policies}. Evolution is performed in a discrete action context with a Genetic Programming approach \citep{koza1994genetic} that only specifies a partial policy as Behavior Trees \citep{colledanchise2018behavior}. An RL algorithm such as \dqn \citep{mnih2015human} or \ppo is then in charge of learning a policy for the states for which an action is not specified. The fitness of individuals is evaluated over their overall behavior combining the BT part and the learned part, but only the BT part is evolved to generate the next generation, benefiting from a Baldwin effect \citep{simpson1953baldwin}. 

Finally, several works consider evolving {\bf the morphology of a mechanical system} whose control is learned with RL. Table~\ref{tab:algos4} only mentions two recent instances, one where the algorithm is not named \citep{gupta2021embodied} and \derl \citep{park2021computational}, but this idea has led to a larger body of works, e.g. \citep{ha2019reinforcement, luck2020data}.

\subsection{Evolution improved with RL mechanisms}
\label{sec:inspi}

Without using a full RL part, a few algorithms augment an evolutionary approach with components taken from RL.

First, the \tres (for Trust Region Evolution Strategies) algorithm \citep{liu2019trust} incorporates into an ES several ideas from the \trpo \citep{schulman2015trust} and \ppo \citep{schulman2017proximal} algorithms, such as introducing an importance sampling mechanism and using a clipped surrogate objective so as to enforce a natural gradient update. Unfortunately, \tres is neither compared to the \nes algorithm \citep{wierstra2014natural} which also enforces a natural gradient update nor to the safe mutation mechanism of \cite{lehman2018safe} which has similar properties.

Second, there are two perspectives about the \zoac algorithm \citep{lei2022zeroth}. One can see it as close to the \zospi algorithm described in Section~\ref{sec:actions}, that is an actor-critic method where gradient descent to update the actor given the critic is replaced by a more robust derivative-free approach. But the more accurate perspective, as put forward by the authors, is that \zoac is an ES method where the standard ES gradient estimator is replaced by a gradient estimator using the advantage function so as to benefit from the capabilities of the temporal difference methods to efficiently deal with the temporal credit assignment problem.

Finally, with their {\sc guided ES} algorithm \citep{maheswaranathan2019guided}, the authors study how a simple ES gradient estimator can be improved by leveraging knowledge of an approximate gradient suffering from bias and variance. Though their study is general, it is natural to apply it to the context where the approximate gradient is a policy gradient, in which case {\sc guided ES} combines evolution and RL. This work is often cited in a very active recent trend which consists in improving the exploration capabilities of ES algorithms by drawing better than Gaussian directions to get a more informative gradient approximator \citep{choromanski2018structured, choromanski2019complexity, zhang2020accelerating, dereventsov2022adaptive}. In particular, the \sges algorithm \citep{liu2020self} leverages both the {\sc guided ES} idea and the improved exploration ideas to produce a competitive ES-based policy search algorithm.

\section{Conclusion}

In this paper we have provided a list of all the algorithms combining evolutionary processes and deep reinforcement learning we could find, irrespective of the publication status of the corresponding papers. Our focus was on the mechanisms and our main contribution was to provide a categorization of these algorithms into several groups of methods, based on the role of evolutionary optimization in the architecture.

We have not covered related fields such as algorithm combining deep RL and imitation learning, though at least one of them also includes evolution \citep{lu2021recruitment}. Besides, we have not covered works which focus on the implementation of evolution and deep RL combinations, such as \citep{lee2020efficient} which shows the importance of asynchronism in such combinations.

Despite these limitations, the scope of the survey was still too broad to enable deeper analyses of the different combination methods or a comparative evaluation of their performance. In the future, we intend to focus separately on the different categories so as to provide these more in-depth analyses and perform comparative evaluation of these algorithms between each other and with respect to state of the art deep RL algorithms, based on a unified benchmark. 

Our focus on elementary mechanisms also suggests the possibility to design new combinations of such mechanisms, that is {\bf combining the combinations}. For instance, one may include into a single architecture the idea of selecting samples sent to the replay buffer so as to maximize the efficiency of the RL component, more efficient crossover or mutation operators as in \pderl, soft policy updates, hyperparameter tuning etc. No doubt that such combinations will emerge in the future if they can result in additional performance gains, despite the additional implementation complexity.

\section*{Acknowledgments}
The author wants to thank Giuseppe Paolo, St\'{e}phane Doncieux and Antonin Raffin for useful remarks about this manuscript as well as several colleagues from ISIR for their questions and remarks about the algorithms.

\bibliographystyle{ACM-Reference-Format}
\bibliography{survey_evorl}

\end{document}